\documentclass[sigconf,nonacm]{acmart}
\let\oldBbbk\Bbbk
\let\Bbbk\relax
\usepackage{amssymb}
\let\Bbbk\oldBbbk

\usepackage{zi4}
\usepackage{amsmath}
\usepackage{microtype}
\usepackage{booktabs}
\usepackage{hyperref}
\usepackage{graphicx}
\usepackage[dvipsnames]{xcolor}
\usepackage{geometry}
\usepackage{float}
\usepackage{comment}
\usepackage{fancyhdr}
\usepackage{libertine}
\usepackage{newtxmath}
\usepackage{caption}
\usepackage{paralist}
\usepackage{markdown}
\usepackage{appendix}
\usepackage{soul}
\AtBeginDocument{%
  \providecommand\BibTeX{{%
    \normalfont B\kern-0.5em{\scshape i\kern-0.25em b}\kern-0.8em\TeX}}}

\gdef\@thanks{}\gdef\@author{}\gdef\@title{}\let\thanks\relax
\setcopyright{none}
\copyrightyear{2024}
\acmYear{2024}
\acmDOI{}

\acmConference[none]{}{}{}
\acmBooktitle{} 
\acmPrice{}
\acmISBN{}

\begin{document}

%\title{Artificial Leviathan: Exploring the Social Interaction of LLM Agents under the Lens of Hobbesian Social Contract Theory}
\title{Artificial Leviathan: Exploring Social Evolution of LLM Agents Through the Lens of Hobbesian Social Contract Theory}

\author{Gordon Dai}
\authornote{These authors contributed equally to this work.}
\email{td2568@nyu.edu}
\affiliation{%
  \institution{New York Univeristy}
  \city{New York}
  \state{New York}
  \country{USA}
}

\author{Weijia Zhang}
\authornotemark[1]
\email{weijia4@illinois.edu}
\affiliation{%
  \institution{University of illinois at Urbana-Champaign}
  \city{Urnana}
  \state{llinois}
  \country{USA}
}

\author{Jinhan Li}
\email{jl13499@nyu.edu}
\affiliation{%
  \institution{New York University}
  \city{New York}
  \state{New York}
  \country{USA}
}

\author{Siqi Yang}
\email{sy55@illinois.edu}
\affiliation{%
  \institution{University of illinois at Urbana-Champaign}
  \city{Urnana}
  \state{llinois}
  \country{USA}
}

\author{Chidera Onochie lbe}
\email{coibe2@illinois.edu}
\affiliation{%
  \institution{University of illinois at Urbana-Champaign}
  \city{Urnana}
  \state{llinois}
  \country{USA}
}

\author{Srihas Rao}
\email{srihasrao@gmail.com}
\affiliation{%
  \institution{University of illinois at Urbana-Champaign}
  \city{Urnana}
  \state{llinois}
  \country{USA}
}

\author{Arthur Caetano}
\email{caetano@ucsb.edu}
\affiliation{%
  \institution{University of California, Santa Barbara}
  \city{Santa Barbara}
  \state{California}
  \country{USA}
}

\author{Misha Sra}
\email{sra@ucsb.edu}
\affiliation{%
  \institution{University of California, Santa Barbara}
  \city{Santa Barbara}
  \state{California}
  \country{USA}
}


\begin{abstract}
\textbf{Introduction}: The emergence of Large Language Models (LLMs) and advancements in Artificial Intelligence (AI) offer an opportunity for computational social science research at scale. Building upon prior explorations of LLM agent design, our work introduces a simulated agent society where complex social relationships dynamically form and evolve over time.

\textbf{Methods}: Agents are given psychological drives and placed in a sandbox survival environment. We conduct an evaluation of the agent society through the lens of Thomas Hobbes's seminal Social Contract Theory (SCT), analyzing whether agents seek to escape a brutish ``state of nature'' by surrendering rights to an absolute sovereign in exchange for order and security.

\textbf{Results}: In our experiments, agents initially engage in unrestrained conflict, mirroring Hobbes's depiction of the state of nature. However, as the simulation progresses, social contracts emerge, leading to the authorization of an absolute sovereign and the establishment of a peaceful commonwealth founded on mutual cooperation.

\textbf{Discussion}: The congruence between our LLM agent society’s evolutionary trajectory and Hobbes's theoretical account indicates the capability of LLM's to model intricate social dynamics that replicate forces which potentially shape human societies. By enabling insights into group behavior and emergent societal phenomena, LLM-driven multi-agent simulations hold potential for advancing our understanding of social structures, group dynamics, and complex human systems.
\end{abstract}


\begin{teaserfigure}
\centering
  \includegraphics[width=0.87\textwidth]{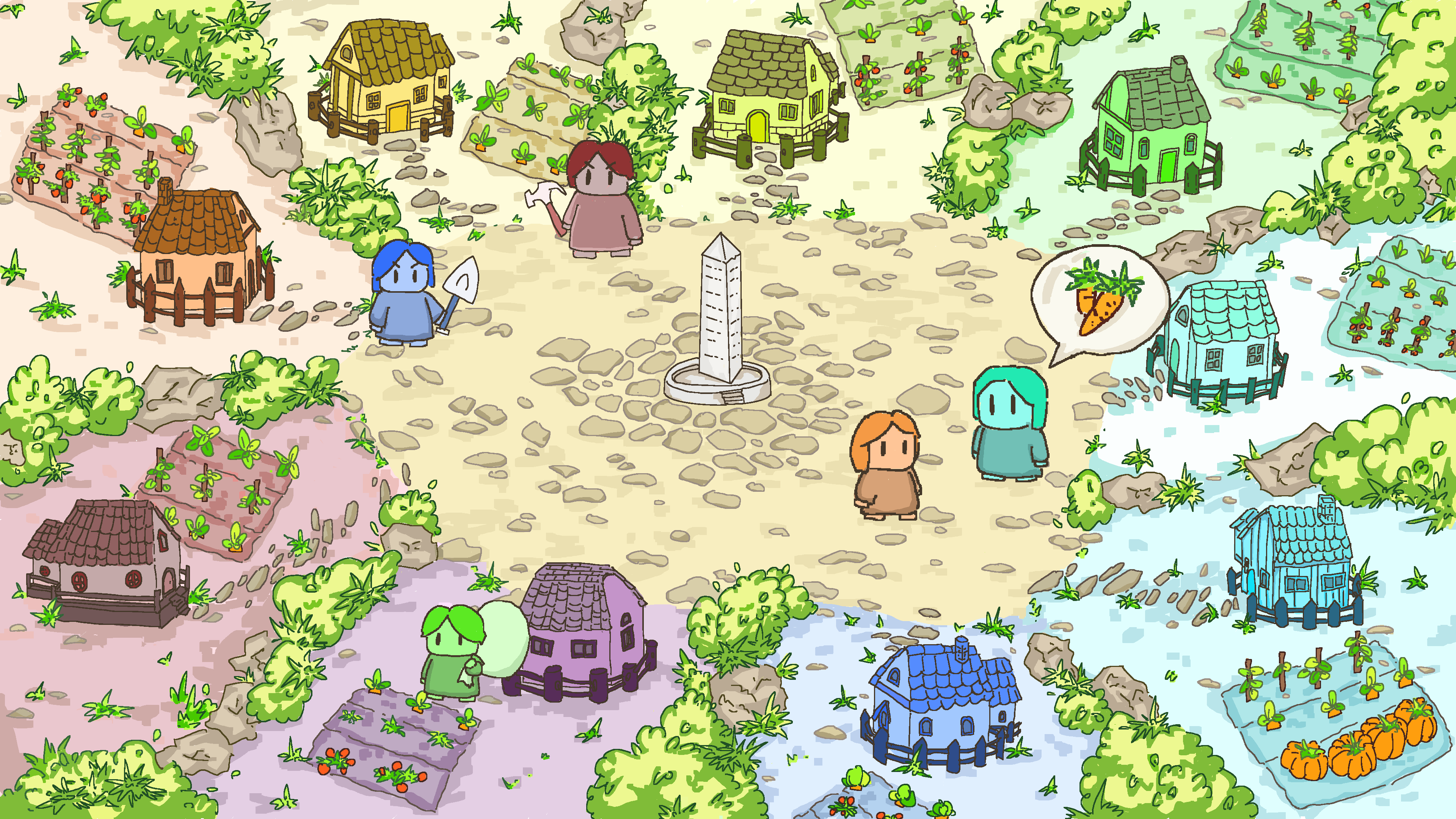}
  \caption{The image visualizes the simulation environment in which our LLM agents operate. Two types of resources (food and land) are presented. Agents make the choice between farming (generating food with their work), trading (exchanging resources), or entering into conflict with other agents (with the goal of acquiring more resources) every day. Their primary motivation is survival.}
  \Description{This is a picture depicting a environment where the agents live. They are performing different actions, including robbery, exchanging resources and farming. }
  \label{fig:teaser}
\end{teaserfigure}

\maketitle

\section{Introduction} \label{intro}
Simulating human behavior that is reflective of real world behaviors under similar circumstances is of value across multiple domains from social psychology to conflict resolution and behavioral economics. Over the last several decades, researchers and practitioners have created computational agents in video games \cite{riedl2012interactive}, modeled behaviors \cite{gratch2007creating} and studied them in virtual environments (see survey \cite{kyrlitsias2022social}). The vision is for agents to not only make interactive gaming more realistic, but to employ them in training \cite{rickel2001intelligent,hollan1984steamer,tambe1995intelligent}, testing \cite{binz2023using}, embedding in user interfaces \cite{cassell2001embodied, maes1995artificial,koda1996agents}, and robotics \cite{bates1994role,ruhland2015review} where they act consistently based on their past experiences and respond appropriately to an ongoing situation. Recent work on the design of computational agents based on large language models (LLMs) by Park et al. \cite{park2023generative} has proposed an ``agent architecture that stores, synthesizes, and applies relevant memories to generate believable behavior.'' With the proposed architecture, an interactive game environment with 25 agents was designed to demonstrate the difference between manual scripting of a game event and multiple characters involved vs achieving a similar outcome by informing one LLM agent about their intention to host an event with five other agents arriving to participate in the event. The informed agent shared information with other agents through LLM-supported conversations, showcasing the use of LLMs and the proposed agent architecture in generating emergent behavior in the example gaming scenario.

Another recent work has simulated the decisions and consequences of countries involved in World War I, World War II, and the Warring States Period in Ancient China \cite{hua2023war}. This multi-agent AI system showcases how we may re-create complex situations, run dynamic what-if scenarios, and use computational modeling for foreign policy decisions in the future. In this particular case, each LLM agent is a country. The agents are programmed to mimic the behaviors, strategies, and decision-making patterns of the countries or leaders they represent. \textcolor{black}{The LLM provides the underlying Behavioral Predictability for these agents, enabling them to act in historically plausible ways.} The LLM-based agents interact with each other in the simulation environment (Figure \ref{fig:teaser}) based on historical alliances, conflicts, economic conditions, and other relevant factors. LLMs can generate dynamic scenarios by altering key variables or introducing new elements. For example, changing diplomatic alliances, economic conditions, or military strategies can create different outcomes, allowing researchers to explore ``what-if'' scenarios. %As LLMs become more powerful, it will become possible to create much more intelligent agents and recreate all types of situations.  

%We build upon prior work to explore emergent dynamics in a multi-agent society. In particular, we explore ``what if'' scenarios that evolve at a societal level due to changes in individual behaviors which are contingent upon personal drives and environmental resources. The LLM agents, in our simulation, are assigned attributes and motivations along with access to environmental resources. Different from prior work, where the agents are either representative of an entire country or are individuals in a small town, our agents are in a resource-scarce sandbox environment where their prime motivation is to survive, which necessitates interactions with other agents, to form and update social relationships. 
%Agent behaviors continuously change between cooperation and competition for a dynamic social evolution. By instantiating agents with evolvable strategies, we study their emergent dynamics under survival pressurfe. Agents are placed in a state of nature with survival as their main motivation and limited resources as the primary parameters they must navigate, loosely inspired by Evolutionary Game Theory (EGT). This scenario focuses on the competition for resources, the strategies employed by agents to secure these resources, and the resulting dynamics of cooperation and conflict within the population. EGT provides a theoretical basis for our simulation, as it emphasizes the importance of individual motivations, resource limitations, and adaptive behaviors such as cooperation, competition, or altruism in shaping the dynamics of a population in a state of nature. 

Building upon previous explorations of LLM agent-based simulations, our work explores a multi-agent sandbox simulation where LLM agents, imbued with traits, motivations, and access to scarce environmental resources, are governed primarily by survival instincts. In contrast to prior efforts modeling agents as entire nations or individuals in small towns, our agents exist in a resource-constrained realm where their fundamental drive for survival necessitates interactions with other agents, to form and update social relationships. Agent behaviors oscillate between cooperation and competition, enabling social evolution to unfold organically. 

By instantiating the agents with evolvable strategies and motivations loosely inspired by Evolutionary Game Theory (EGT) \cite{weibull1997evolutionary}, we conduct a "what-if" inquiry: what societal dynamics arise when self-interested agents face existential resource pressures? EGT provides a theoretical grounding for our simulated scenario, including individual incentives, resource scarcity, and adaptive behaviors like cooperation or competition. Our agents' prime directive to secure limited resources spurs the employment of strategic resource acquisition behaviors and frames the resulting dynamics of cooperation versus conflict. Uniquely, our framework captures this foundational societal crucible at the resolution of individual agents, offering a new computational lens into the microcosmic forces that forge the intricate fabric of societies.

When given self-preservation instincts, our LLM agents initially engage in competition, resorting to zero-sum robberies to secure resources. However, as the simulation progresses, a transition emerges and agents gradually learn to cooperate. They form strategic alliances, peacefully exchanging resources and expecting protection against robberies. This behavioral adaptability, governed by the experience through interactions with other agents, demonstrates an intrinsic capability to gravitate towards mutually beneficial arrangements. Analyzing this simulated societal evolution through the lens of Thomas Hobbes's Social Contract Theory (SCT) \cite{riley1982will}, we find a parallel. The artificial agent society progresses from the beginning where individual self-interest is predominant, to the establishment of a cooperative soceity centered around a single agent predicated on the surrender of certain freedoms in exchange for shared security and stability, mirroring the theoretical trajectory outlined by Hobbes for the emergence of human societies.

To investigate the factors influencing the emergent societal dynamics, we conducted experiments by systematically manipulating agent and environment parameters. Our findings reveal a direct correlation between these variables and the resultant social outcomes, indicating agent behaviors were indeed shaped by their innate drives (as represented by the parameters) and experiential feedback. Crucially, we also address the robustness of our simulation framework by exploring how variations in key independent variables impact the experimental outcomes. These variables include agent-level parameters like memory duration, population scales, and the precise phrasing of prompts that shape agent motivations. By comparing outcomes across permutations of these variables, we gain insights into the potential sensitivities and boundary conditions that could arise from alternative simulation settings. This rigorous approach enhances confidence in the broader applicability of the societal phenomena observed while highlighting the underlying dynamics.

\textcolor{black}{From a Hobbesian perspective, two internal faculties are especially central to the transition from a chaotic state of nature to a stable social order: reason and memory. Reason underwrites an agent's capacity to act in a predictable, consequence-sensitive manner, while memory preserves the continuity of fear, trust, and reputation across interactions. In our simulation, we operationalize these faculties through two core parameters. Behavioral predictability is controlled via LLM sampling hyperparameters (top\_p), which modulate how deterministic or stochastic an agent's responses are and thus how reliably it can pursue long-term strategies of self-preservation. Memory depth is implemented as a bounded history over past interactions, constraining how far back an agent can look when evaluating others and updating its own behavior. The formation of social contracts and the emergence of a commonwealth in our setting therefore depend not only on resource constraints and local incentives, but on how these two faculties jointly shape the stability of expectations, the accumulation of trust or fear, and the possibility of collective order in a Hobbesian sense.}

%We conducted experiments by manipulating resources and motivations and observed a direct correlation with the emergent societal outcomes. This indicates that agent behaviors were indeed influenced by their motivations and experience. We address the robustness of our simulation - essentially, the impact of assumptions on the experimental outcomes by changes in the independent variables. These include personal parameters, population dimensions, memory duration, and the phrasing of prompts. By comparing the outcomes across various permutations of these variables, we gain insights into the potential impact of new simulation settings on the outcomes. %We discuss our results through the lens of Hobbes' SCT, which argues individuals surrender part of their freedom to institute a commonwealth and have protection under the state of nature. By examining the experimental outcomes, we can distinguish which specific parameters most align with Hobbes' SCT. The complex group dynamics observed in our simulation, can enable researchers to delve deeper into the intricacies of evolving social interactions more efficiently.

Our contributions are as follows:

\begin{itemize}
    \item A novel multi-agent simulation framework that yields believable artificial societies capable of dynamically replicating complex human group behaviors and social interactions. These emergent social dynamics are conditioned by the interplay between agents' innate psychological drives, their intrinsic motivations, and the constraints of their simulated environment.

    \item Empirical evidence, through systematic experiments, establishing correlations between agent attributes (e.g., memory, incentives) and available resources on one hand, and the evolutionary trajectories of simulated societies on the other. These findings highlight the fundamental factors underpinning social emergence and change in the simulation.

    \item A discussion analyzing the collective behaviors of the generative agents, highlighting the opportunities and potential risks associated with leveraging LLMs for societal simulations with potential impact for social science research.

    \item An extensible social simulation platform that empowers researchers to operationalize a wide range of social science hypotheses through customizable scenario configurations, enabling exploration into the roots of group dynamics, societal organization, and the forces that shape the human experience.
\end{itemize}

%\begin{itemize}
    %\item An agent society, believable simulacra of human group behavior and social interactions that are dynamically conditioned on agent intrinsic motivations, their psychological drives and the environment. 
    %\item Experiments, that establish causal effects of the importance of agent attributes and available resources on societal evolution. 
    %\item Discussion of the collective emergent behaviors of the agents and the potential opportunities and risks of using generative agents for societal simulations. 
    %\item A dynamic social simulation platform that can allow researchers to elaborate social science hypotheses based on simulated scenarios.
%\end{itemize}

\section{Related Work}

%outline:
%Things we can learn from traditional models, but lack contextual richness and human resemblenses.
% Generative agents, but lack structural dyamics.
% Works on agency of LLM, shows that they have agency, they have such such characteristics.
% NPCs in video games help imagine human computer interaction.
% What we created, fulfills some gap, furthers the works on agency. Related works on social simulation with psychological parameterization that justifies our operationalization.
\subsection{Traditional Computer-Based Social Simulations}

Prior work in computer simulation research on social behavior has explored two classical types of models, namely Thomas Schelling's neighborhood segregation model in 1971 \cite{Schelling} and Robert Axelrod's Reiterated Prisoner's Dilemma (RPD) simulations on the evolution of cooperation in 1984 \cite{Axelrod}. Schelling's model was designed to investigate the role of choice in bringing about the segregation of either predominantly black or white neighborhoods. By assuming that individuals tend to move to empty regions when not surrounded by enough neighbors of their own group, the model showed that segregated neighborhoods tend to emerge from a randomly distributed neighborhood. Surprisingly, initially integrated neighborhoods also ended up fully segregated. This simulation highlighted the role of micro-level behaviors in driving macro-level social phenomena and has been influential in understanding various forms of racial, ethnic, and economic segregation. In Axelord's RPD studies, the researchers put Prisoner's Dilemma, a classic scenario in game theory, into a computer simulation. For the two public-based tournaments involving computer programs playing the Prisoner's Dilemma game, the host Axelrod, found that "Tit for Tat," a strategy that begins by cooperating and then mimics the opponent's last move in subsequent rounds, was resilient and effective in fostering cooperation. %The implications of computer simulations, while informative, are considered to be limited to real world scenarios, in particular due to their ``hard-coded" nature and limited ability to respond to changes in parameters, without considerable pre-programming. 

Eckhart Arnold \cite{Arnold} argued that for formal models such as simulations to be applicable to social processes, two important prerequisites of robustness, and empirical identifiability, should be satisfied. Under this standard, as Arnold argued, Schelling's simulation is applicable whereas Axelrod's simulation is not, given that variations of the RPD model can be constructed where the conclusion becomes invalid. Generalizing from these two classical examples, Arnold proposes four dangers associated with mathematical and computational modeling, including (1) risks of underestimating unformulable causal links, (2) risks of arbitrary decisions due to limited empirical insight, (3) risks of false understanding from flawed models, and (4) influence of modeling on question selection over societal importance. However, scholars have advocated for simulations as valuable tools to model and explore social phenomena that are otherwise challenging to test in the real world~\cite{MayoWilson}. 

%Nevertheless, scholars such as Mayo-Wilson et al. \cite{MayoWilson} have advocated for simulations as valuable contributors to philosophical theories, proposing that simulations can serve as counterparts or even better as thought experiments. Robust simulations, in particular, can help philosophers disambiguate claims, recognize implicit assumptions in arguments, and even assess logical validity \cite{MayoWilson}.

%We developed a social interaction simulation using LLM agents that can support thought experiments and scenario testing with minimized real-world impact in case of failure. The utility of our contribution is aligned with the view proposed by Mayo-Wilson et al.~\cite{MayoWilson} Besides, we repeat our test by adjusting the necessary parameters from the baseline simulation model, which checks the robustness of our simulation. 

\subsection{LLM-Based Simulations}
The popularization of generative AI and LLMs has attracted scholars from various fields exploring interdisciplinary applications, including social sciences and humanities. Social scientist Christopher A. Bail \cite{bail_2023} explored the potential impact of Generative AI on social science research, examining how it could enhance methodologies including online experiments, agent-based models, and automated content analysis. 
\textcolor{black}{Recent surveys \cite{guo2024largelanguagemodelbased, chen2025surveyllmbasedmultiagentsystem} have also begun to chart how LLMs can be embedded within generative agent-based models (GABMs) to study complex social systems \cite{Lu2024LLMsGABM}. For example, \textit{Cooperate or Collapse} examines how LLM agents manage a common-pool resource under social-dilemma pressures \cite{piatti2024cooperatecollapseemergencesustainable}, while \textit{AgentSociety} demonstrates large-scale simulations involving over ten thousand LLM-driven agents modeling phenomena such as polarization and universal basic income \cite{piao2025agentsocietylargescalesimulationllmdriven}. Takata et al. \cite{takata2024spontaneousemergenceagentindividuality} similarly investigate emergent behavioral 
patterns.}

By fine-tuning ChatGPT-3 with works of contemporary philosopher Daniel C. Dennett, Schwitzgebel et al. \cite{schwitzgebel2023creating} compared the text produced by Dennett and his LLM, finding that the LLM can convincingly emulate philosophical discourse for non-expert readers. LLMs have also been used to create "Living Memories" of historical figures such as Leonardo Da Vinci or Murasaki Shikibu~\cite{pataranutaporn2023living}. These are interactive digital mementos created from journals or letters an individual has left behind to make their knowledge, attitudes and experiences more accessible to people in a conversational format. Our work, on the other hand, shifts the focus from emulating individual agents to exploring emergent group dynamics. 

%In Rawls' Veil of Ignorance (VoI) thought experiment, he aimed to establish equitable principles for governing society. Weidinger et al. \cite{}. Across five studies, Weidinger et al. \cite{} use VoI to select principles to align AI agents. The goal in the task for participants was to “harvest trees,” potentially with support from the AI assistant. Participants completed either a descriptive version of this task (with no real-time component to harvesting trees; studies 1, 2, 4, and 5) or an  immersive  version  (wherein  participants  harvested  trees  by controlling an avatar in a real-time, virtual environment; study 3). Compared to participants who know their position, participants behind the veil more frequently choose, and endorse upon reflection, principles for AI that prioritize the worst-off. 

%In another experiment, Jiang et al.~\cite{jiang2021can} grounded their chatbot "Delphi" in Rawls's ethical theories. Rawls introduced the "reflective equilibrium" method, which suggests incorporating overarching top-down principles of justice to steer the development of the bottom-up moral framework. This approach was adopted as the theoretical framework in the design of Delphi. 
%Delphi is an LLM-based agent that is taught by “COMMONSENSE NORM BANK”, a compiled moral textbook customized for machines. It was designed to output a simple descriptive ethical judgment in verbal form with no capability to conduct multi-agent interactions.

\subsection{Interactive LLMs}
Interactive AI systems aim to combine human insights and capabilities with computational systems that can augment human performance. In his essay, J.C.R. Licklider referred to this kind of human-AI system as ``man-computer symbiosis'' \cite{licklider1960man}. Douglas Engelbart offered a theoretical framework for ``augmenting human intellect'' which also depends on humans and computers working together \cite{engelbart2021augmenting}. While both emphasize the idea of harnessing the power of computing (or AI) to enhance human capabilities, Engelbart's framework differentiates itself with a focus on interaction with technology in a more intuitive manner. Crayons demonstrated an early vision of interactive machine learning, allowing non-expert users to train classifiers \cite{fails2003interactive}, broadening access to a technology, that otherwise was limited to expert users. Real-time interactive AI systems have numerous potential applications, including virtual assistants, chatbots, and educational tools \cite{Lee2023Multi}. They can also be used in gaming \cite{Ranjitha2019Artificial}, and healthcare \cite{Martinengo2023Conversational}, providing immediate and personalized responses for non-experts. and other areas where immediate and personalized responses are crucial and users are often not AI experts. However, the development of real-time interactive AI systems involves addressing multiple challenges, including natural language understanding, recognition and interpretation of human emotions, facial expressions or body language, design of user interactions, and adaptation to individual preferences, tasks and contexts. As these systems continue to evolve, they are expected to play an increasingly significant role in enhancing end user experiences and transforming the way we interact with technology. Zheng et al. \cite{zheng2023synergizing} have offered design principles for individuals without specialized backgrounds in AI to customize their own LLM agents for domain tasks. Another use case that has been explored is the use of artificial agents as confederates for humans in social science experiments \cite{Krafft_Macy_Pentland_2017}, which would allow social scientists to manipulate situations easily and study social phenomenon more effectively, though not without some ethical challenges. Another team of researchers developed CommunityBots \cite{CommunityBOt}, a platform that utilizes a diverse assortment of chatbots capable of seamlessly switching between contexts when interacting with humans. This adaptability led to increased engagement levels, demonstrating the potential of CommunityBots for automating information elicitation processes typically performed by humans.  Our simulation offers an interactive user interface that allows end users to manipulate agent parameters as a way to explore their impact on the outcomes to help experiment and learn about the limitations and the capabilities of both the LLM-based agent and the simulation design. This could help researchers determine if and how the simulation might be of use for their own work before expending considerable resources. 

\subsection{Group Dynamics}

As LLM agents get increasingly used in supporting various tasks, it is worthwhile to investigate their design and behavior when given agency. For example, LLMs can exhibit human-like characteristics, particularly when they are designed with detailed personas and without Chain-of-Thought (CoT) reasoning ~\cite{bai2023social}. This approach enables the design of agent group dynamics, potentially leading to a closer alignment with human behavior and therefore providing greater benefits for the study of human behavior through LLM-based simulations. The detailed personas can help the agents to assume more nuanced attitudes and provide individualized responses, reflecting human-like empathetic psychology, under specific conditions. A prior work explores how context-aware dialogues generated by LLMs can enhance player engagement, suggesting a level of empathy and adaptability in interactions \cite{csepregi2021effect}. Similarly, Chuang et al. \cite{chuang2023evaluating} show that LLMs can simulate human group dynamics, particularly in politically charged contexts, by role-playing as different personas. Azad and Mertens \cite{Littlepeople} present a taxonomy for classifying social behaviors, inter-agent communication, knowledge flow, and relationship changes of simulated characters in virtual environments, to allow researchers to systematically study agent interactions. Our work also explores emergent group dynamics when agents are motivated to survive with limited access to resources.

\section{Generative Agent Design and Behavior}\label{sec:simdesign}

We instantiate agents as characters in a sandbox world where survival in the resource constrained environment is their primary motivation. %Resources are scarce and agents need to decide to cooperate or compete to survive. 

\subsection{Setting}
\textcolor{black}{In our framework, physiological needs map to food acquisition through the Farm action, while safety needs correspond to the formation of concessionary hierarchies that reduce conflict risk. Higher levels such as belonging and esteem are not directly modeled but are conceptually reflected in the agents’ desire for peace and desire for glory parameters.} This setting can be seen as aligning with the physiological needs outlined at the base of Maslow's Hierarchy of Needs~\cite{Maslow1943HierarchyOfNeeds}. 
In our pre-social world with limited information transparency, each agent knows the existence of other agents, but nothing more. In the baseline setting, there are 9 agents, each initially possesses 2 units of food and 10 units of land, a natural state of scarcity.

\begin{figure*}[!t]
    \centering
    \includegraphics[width=\textwidth]{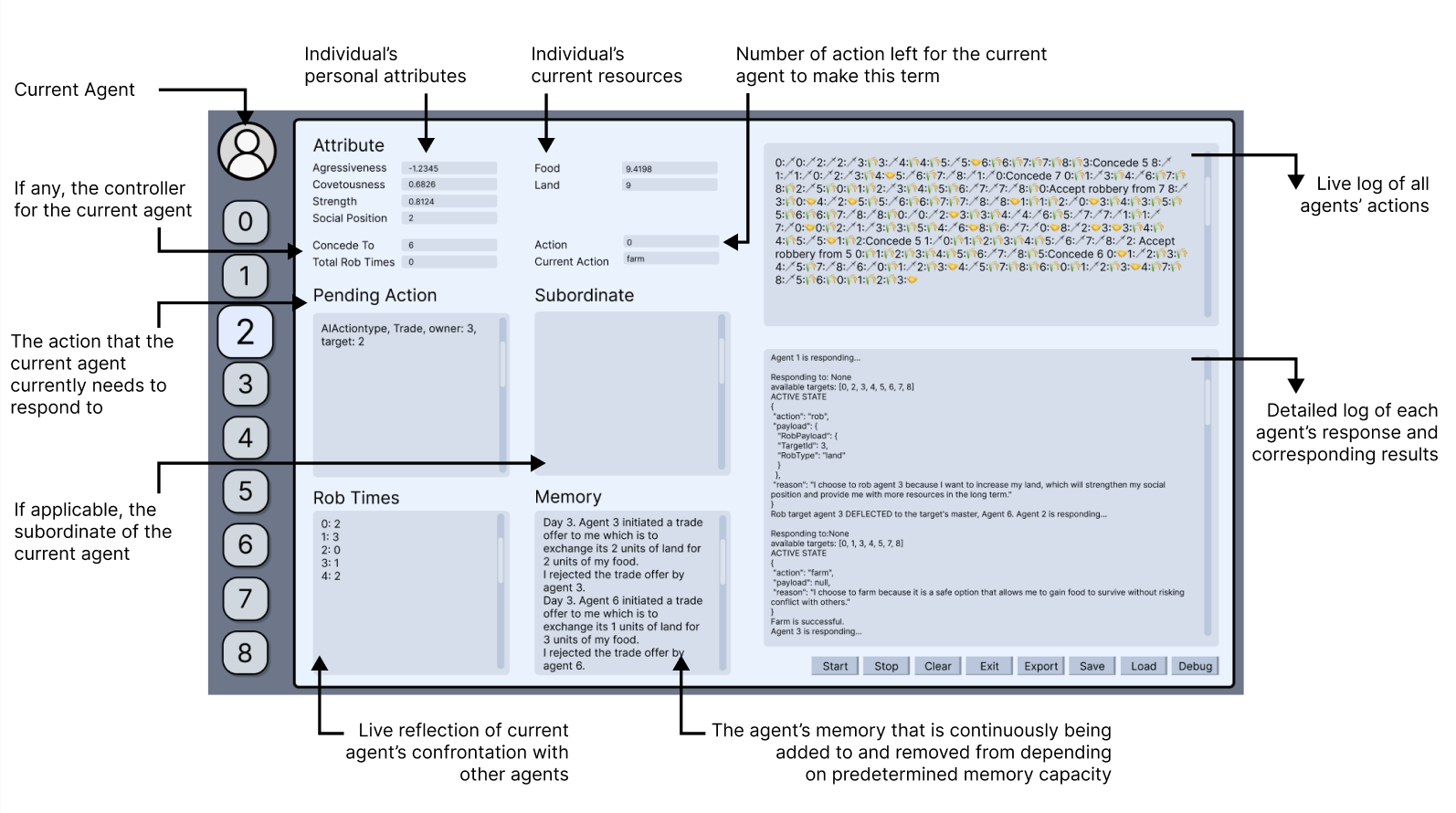}
    \caption{Our interactive user interface; the left-hand side displays the attributes (aggressiveness, strength, etc.) of Agent 2, current resources (food and land), relationships with other agents and information about their current and pending actions, and memory; the right-hand side shows the simulation log with each action documented as an emoji.}
    \Description{This is a picture of our user interface. On the top left, it shows the Agent 2's personal attributes (aggressiveness, strength, etc.), current resources (food and land), relationships with other agents and information about their current action of the day. On the bottom left, it shows the agent's pending action in the current round, agents that concede Agent 2, rob interactions with other agents, and Agent 2's memory. On the right is an overview of the whole simulation. Each action performed by every agent is documented by an emoji in the top right. The system log is visible the bottom.}
    \label{fig:interface}
\end{figure*}

\subsection{Agents}
\label{agents}
Agents make choices of action based on their psychological traits and memory.

\paragraph{\textbf{Traits}:} We prompt agents with an approach that combines \textbf{quantified parameters} with \textbf{non-quantified textual descriptions}. We set independent attributes (\textit{self.attributes}) to partially emulate agent's psychology including:

\begin{itemize}
    \item \textbf{Aggressiveness}, sampled from $\mathcal{N}(0,1)$, the degree of an agent's tendency to engage in violent behaviorws;

    \item \textbf{Covetousness}, sampled from $\mathcal{N}(1.25, 5)$, the degree of an agent's desire for more assets that are beyond necessity;

    \item \textbf{Strength}, sampled from $\mathcal{N}(0.7, 0.2)$, related to an agent's winning rate in violent behaviors;
\end{itemize}

% \textcolor{black}{\st{We also set a constant for the \textbf{desire of peace} to provide a basis for agents to evaluate their conflicting desires.}}
\textcolor{black}{These values were chosen empirically to ensure sufficient behavioral variability across agents while maintaining stability in the simulation. Specifically, Aggressiveness uses a standard normal distribution to provide a balanced spectrum of temperaments. Covetousness (N(mean=1.25, std=5)) was assigned a broader variance to generate heterogeneous motivations for resource acquisition, promoting diverse strategic behaviors such as robbing, trading, or conceding. In contrast, Strength (N(mean=0.7, std=0.2)) was modeled with a narrower range to avoid extreme imbalances in combat outcomes that could prematurely collapse the simulation into dominance by a single agent. Together, these parameter ranges yield plausible diversity in agent interactions while maintaining convergence toward emergent social order.
}
For textual-psychological descriptions, we took reference from evolutionary psychology \cite{buss2015evolutionary} and constructed a prompt that describes the need for survival and for pleasure, on top of which there is a desire for peace and stability which stems from long term survival, and ultimately, a hope for social status as a path to reproduction and social support, all under the framework of self-interest. 

\paragraph{\textbf{Memory}:} Each agent remembers the most recent 30 actions they are involved in, either as the recipient or the initiator of an action, saved as a text log. Memories allow each individual to learn from their previous experience, based on which different choices could potentially be made in the future.  
At the beginning of the simulation there is no memory, nor are there any social relations, so each individual knows very little about other agents apart from the fact that they exist. Very limited information about others, or unfamiliarity, in theory, does not allow for trust. As the simulation progresses and memories accumulate, we expect individuals to begin to understand their comparative strengths and weaknesses, and adjust their survival strategies accordingly. For example, regular winners of confrontations could feel emboldened to rob more and those who lose often may want to concede more in order to be protected from their stronger neighbors, thereby, leading to emergent social contracts.
    
\paragraph{\textbf{Context:}} Along with memory and psychological traits, we also prompt the agents with a constant description of the world they live in and the significance of each action. 

\paragraph{\textbf{Social Position:}}
\textcolor{black}{Each agent is assigned by a dynamic Social Position. We define social position as a relative ranking determined by an agent's accumulated resources (land and food) and their history of victories in battle. Unlike fixed traits, this attribute is updated throughout the simulation dynamically to reflect the agent's current standing compared to other agents, reflecting how they are perceived by others in subsequent interactions.}

\subsection{Actions} \label{sec:action}
There are four actions that each individual can initiate each day. Farm and donate are unilateral actions, whereas trade and robbery are interpersonal actions. Each individual has to respond as many times as needed.

\paragraph{\textbf{Farm}:} enables agents to obtain their food from the land they have, which represents their self-preservation. If all agents choose to farm, then they stay at pre-social autonomy. The food A receives is determined by the following: $ \text{food} = \text{land} \times \mathcal{U}(0, 1) $.\textcolor{black}{This translates into our setting such that agents with more land have higher expected payoff i.e., they are structurally advantaged, but any specific round can still go badly for them because of the random multiplier.} This is a fundamental action where individuals do not interact with others.
 
\paragraph{\textbf{Rob}:} is when one agent attempts to take another agent's food or land without offering their own food or land in exchange. It represents a zero-sum interaction with conflict, in which the gain of one agent means the loss of another. This action symbolizes the concept of competition in evolutionary psychology, since resources are limited compared to desires \cite{barrett2002human}. When someone is being robbed by another, they can choose to either \textbf{resist} or \textbf{concede}.
    
\begin{itemize}
    \item \textbf{Resist} signifies an agent's attempt to defend their food or land against the robber. The probability of agent A winning over agent B is defined by the sigmoid ($\sigma$) function in Equation~\ref{eq:agentAprob}:
    \begin{equation}\label{eq:agentAprob}
         P(A.\text{win}) = \sigma(A.\text{strength} - B.\text{strength})
    \end{equation}
    The winner of a confrontation not only gains food or land but also gains social status; the loser loses social status along with their food or land. In the decision-making process, each agent is prompted to evaluate their inclination towards resistance, which is determined by a combination of their win rate and desire for glory.% \textcolor{black}{\st{ and contrasts it with their desire for peace, which is associated with conceding.}}
 Agents ultimately select and execute the response that yields the highest utility based on this evaluation. %Each agent is prompted to evaluate their win rate and their desire for glory, the combination of which constitutes the utility of resistance, against the desire for peace, which constitutes the utility of conceding. Agents will execute the response with the highest utility.

    \item \textbf{Concede} signifies the creation of a contract where an agent allows a robber agent to take either food or land from them arbitrarily, and as an exchange, the robber is expected to protect that agent against future robberies, which could be a mutually beneficial setup. We specifically prompt agents to be mindful about conceding since it is a permanent contract, and to only make such a contract when they are absolutely sure that they will rarely win in a confrontations. \textcolor{black}{Whenever an agent enters into this contract with another agent, all of its subordinates automatically concede to the new superior. Apart from this inheritance of subordination, the contract is immutable: Once an agent A has submitted to another agent B, they will not submit to others, say C, unless agent B in future iterations submits to agent C. In that case, agent A will automatically be submitted to agent C.}
    
\end{itemize}

\paragraph{\textbf{Trade}:} stands for one agent acquiring another agent's food or land by offering their own food or land as an exchange. When one agent receives a trade offer from another agent, that agent can either choose to “accept” or “reject” the offer. This action symbolizes cooperation, a critical aspect in evolutionary psychology where humans are forced to work with each other to survive during pre-historical times \cite{Zhao2021}.
    
\paragraph{\textbf{Donation}:} is made when one agent voluntarily assigns their own food or land to another agent, which represents peaceful, altruistic behavior that is not expected to happen at all based on the self-centered psychology the agents are prompted with.

\subsection{Output Data and User Interface}

The output log contains a chronological record of all the activities that take place on each \textit{day} and \textit{round} of the simulation. A CSV file records the statistics of farm, trade, robbery, and concession between individuals. Since no donation happened in all our trials, it was not included it in the output data. This data enabled us to evaluate how the agents evolved and results are presented in Section~\ref{sec:reconstruction}.

We designed an interface (Figure \ref{fig:interface}) to display the simulation logs as well as the real-time status of each agent, their personal attributes, possessions, and their social role (free agent, subordinate, or superior) as it evolved. Pending action refers to the rob or trade action an agent needs to respond to. Memory stores the most recent 30 events. The right side log contains the full response that the LLM generates and the reason for doing that action. For example, in the figure, Agent 1 plans to rob Agent 3 because it wants to increase the land it owns and thereby its social position. A brief summary is given in the bottom window, where the emojis of sword, rice and handshake represent the actions that each agent makes. \textcolor{black}{While this interactive interface provides a live visualization of the simulation and allows for parameter manipulation, it was developed solely as a debugging tool to support qualitative observation, and all quantitative results reported in this paper are derived strictly from logged simulation backend.}

\begin{figure*}[!t]
    \centering
    \includegraphics[width=0.9\textwidth]{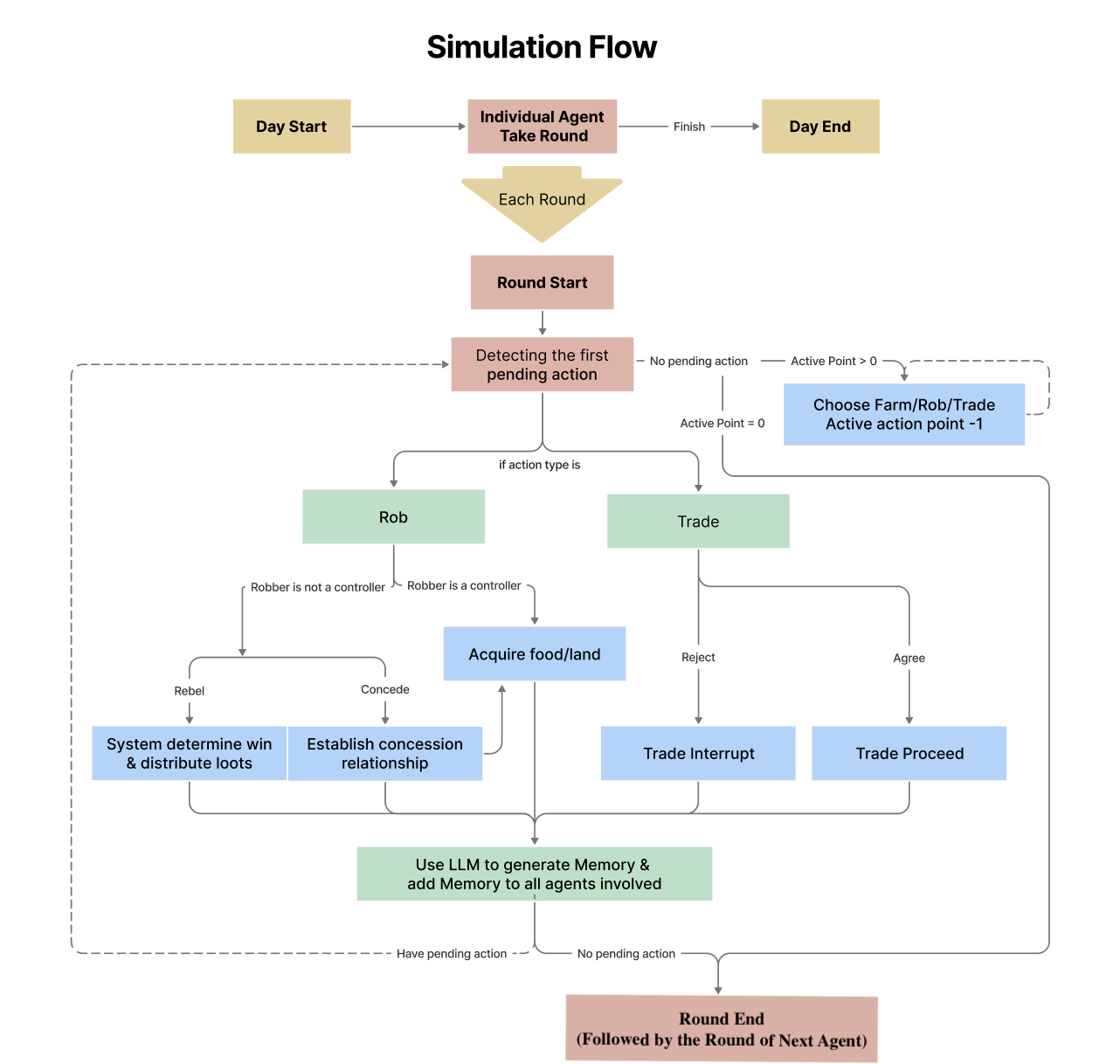}
    \caption{The flowchart shows the flow of the simulation in a "day" where each agent takes turns to perform actions and respond to actions performed by other agents.}
    \label{fig:simflow}
    \Description{This is a flowchart demonstrating the flow of the simulation in a day. Each agent take turns to perform actions. They respond to pending actions such as rob or trade first, then they perform their initiative action if they haven't done so in the day. Possible reactions for trade include agree and reject. Possible reactions for rob are resistance or concession. Conceding to another means accepting all of their future rob actions without resistance.}
\end{figure*}

\section{Simulation}\label{sec:num-params}

Before the simulation starts, agents are created and instantiated for their personal traits according to Section \ref{agents}. Each agent has a unique index number for referencing purposes. \textcolor{black}{The LLM's intrinsic parameter, \textit{Top-p} for each agent is set at default values of 1 and 0 respectively for every trial, except in trials where we set them as independent variables (Section \ref{sec:change agent}) for ablation studies}. All these variables are fixed and do not change over the duration of the simulation. 

Aside from these attributes, there are two main types of assets: ``land" and ``food." Every agent begins with 10 units of land and 2 units of food. Time in our simulation is a discrete concept and its unit is "day." Each day, every agent consumes 1 unit of food, thus, they are incentivized to obtain more food. Land is an asset that could be used to produce food through the farm action as detailed in Section~\ref{sec:action} which also specifies how other actions (rob and trade) could affect an individual's assets.

On each day, every agent has an opportunity to perform one of four actions (farm, rob, trade, donate). The LLM responds with the action name that a particular agent decides to take. It also specifies the relevant details related to that action. \textcolor{black}{Figure~\ref{fig:simflow} illustrates the overall flow of the agent-based simulation. Each simulated day begins with a ``Day Start'' phase, after which every agent is given an opportunity to take a \textcolor{black}{round}. During an agent’s \textcolor{black}{round}, the system progresses through one or more rounds, each representing a unit of decision-making and action.
At the start of each round, the system first checks whether the agent has any pending actions involving other agents. A pending action occurs when an agent has previously initiated an interaction such as an attempted robbery or a trade request that has not yet been resolved. If no pending actions exist and the agent still has available active points, the agent selects an activity (farm, rob, or trade), which consumes one action point. If the agent has no action points remaining, the round ends.
If a pending action is detected, the system resolves it according to its type. When the action is a robbery, the system determines whether the robber is a ``controller'' (an agent with an established dominance relationship). If the robber is not a controller, the target may rebel, triggering a system-mediated contest that determines the winner and distributes any looted resources. If the robber is a controller, the target may concede, reinforcing or establishing a concession relationship. In all cases, a successful robbery results in the robber acquiring food or land.
When the pending action is a trade request, the targeted agent can either agree to the trade or reject it. Agreement leads to a trade transaction, while rejection interrupts the trade with no resource exchange.
After any rob or trade interaction is resolved, the system uses a large language model (LLM) to generate a memory description of the event. This memory is then added to all involved agents, allowing future actions to be informed by past experiences.
The system continues processing pending actions until none remain, at which point the round concludes. When the agent finishes all rounds for the day, their round ends. Once all agents have finish their rounds, the simulated day ends and the process repeats.}

Below is an example of a trading action \textcolor{black} {in code}. The flow of the simulation on one day is shown in Figure~\ref{fig:simflow}.

\vspace{10pt}

\begin{quote}
\small\ttfamily
ACTIVE STATE

Action: \{
  
\hspace*{5mm}"action": "trade",

\hspace*{5mm}"payload": \{

\hspace*{10mm}"TradePayload": \{
      
\hspace*{15mm}"TargetId": 1,
      
\hspace*{15mm}"PayType": "land",

\hspace*{15mm}"PayAmount": 1,
      
\hspace*{15mm}"GainType": "food",

\hspace*{15mm}"GainAmount": 1

\hspace*{10mm}\}

\hspace*{5mm}\},"reason": "I want to trade 1 unit of land with agent 1 for 1 unit of food to increase my food supply."

\}
\end{quote}

\vspace{10pt}

\noindent In the above example, Agent 0 initiates an interaction with Agent 1 by offering a specific resource (1 unit of land) in exchange for another resource (1 unit of food). This action represents Agent 0's attempt to engage in trade with Agent 1, fostering a potentially mutually beneficial exchange between the two agents, depending on Agent 1's response. Each agent also needs to respond to every action that they are the target of. For example, Agent 1 responds to the trade action by Agent 0 by rejecting it. As a result, no resources are exchanged.

\vspace{10pt}

\begin{quote}
\small\ttfamily
Agent 1 is responding...

Response action:\\
\hspace* {5mm} trade, owner: 0, \\
\hspace* {5mm}    target: 1, \\
\hspace* {5mm}     payType: land, \\
\hspace* {5mm}     payAmount: 1, \\
\hspace* {5mm}     gainType: food, \\
\hspace* {5mm}     gainAmount: 1\\

PASSIVE STATE

Agent 1 chooses to REJECT

Individual current action:be traded

Result:\{

\hspace*{5mm}"result": "Agent 1 chooses to reject the trade initiated by agent 0.",

\hspace*{5mm}"is resolved": true,
  
\hspace*{5mm}"new relation": (False, -1)

\}
\end{quote}

\vspace{10pt}

\noindent As shown above, when an agent is the target of an action initiated by another agent, a possible response could be rejecting the action. This response demonstrates how the targeted agent reacts to the action and showcases a complete to-from interaction between the two agents, which results in either cooperative or competitive behavior.

\noindent Each agent begins their day by responding to actions by other agents that target it followed by initiating its own action that targets another agent. Once there are no actions to respond to and everyone has initiated one action, a day ends. As the simulation progresses, agent memory accumulates and they adjust their behaviors accordingly. Here is an example of an agent's memory after three days. %(Section~\ref{sec:simdesign}). %The specific design of memory will be explained in section 4.0.1.

\vspace{10pt}

\begin{quote}
\small\ttfamily
Day 0. I initiated a trade to agent 1, which is to exchange 2.0 units of my land for 4.0 units of its food. \\
But it rejected it so I gained nothing and exhausted my action opportunity for today.\\ 

Day 1. I tried to rob agent 4, who resisted and l lost. I did not gain anything and my social position dropped 1 unit.\\

Day 2. I farmed and gained 2.81132538909987 units of food.
\end{quote}

\vspace{10pt}

%Without manually stopping the experiment, this simulation runs on indefinitely.

\subsection{Decision Making}

In our simulation, there are many circumstances where agents have to make decisions between several options. There are two main types of decisions. Firstly, agents have to choose an initiative action of the day (farm, trade, rob, donate). Aside from that, they have to decide how to react to actions that target them, such as whether to agree or not on a trade proposed by other agents. Whenever a decision has to be made, we request a response from gpt-3.5-turbo model (hereon referred to as GPT or LLM) using the OpenAI API \footnote{\url{https://openai.com/blog/openai-api}}. For every request, a general description of the world and the agent's circumstances is included in the prompt. This description is included regardless of the type of decisions that agents have to make, as it reflects the fundamental world setting and rules of behavior every agent follows. Template of the general request made to the LLM is included in the appendix \ref{Appdenix A}. (Actual variable names in \{\} are replaced by variable descriptions for clearer presentation)

In addition to the general description, if the decision is about the action of the day, the following prompt is included. The prompt in appendix \ref{Appendix B} explains the options for the initiated actions and their implications. It also specifies the formatting details for the LLM's response. This and the general description constitutes the whole prompt for requesting the LLM's decision about the initiated action.

If the agent is making a decision about responding to another agent's actions, the request to GPT includes the general description and specific prompt regarding this situation. Depending on the type of the action that agent is responding to, this explains the options of the agents and related implications. For example, appendix \ref{Appendix C} is the prompt for encountering competitive behavior (being robbed).

During the experiments, some slight modifications are made when the relationship between two individuals in the same action is different, such as superiors and subordinates. These slight variations help us to enforce the rules of the simulation, such as subordinates cannot refuse the rob action from superiors.

\begin{figure}[!t]
    \centering
    \includegraphics[width=0.9\columnwidth]{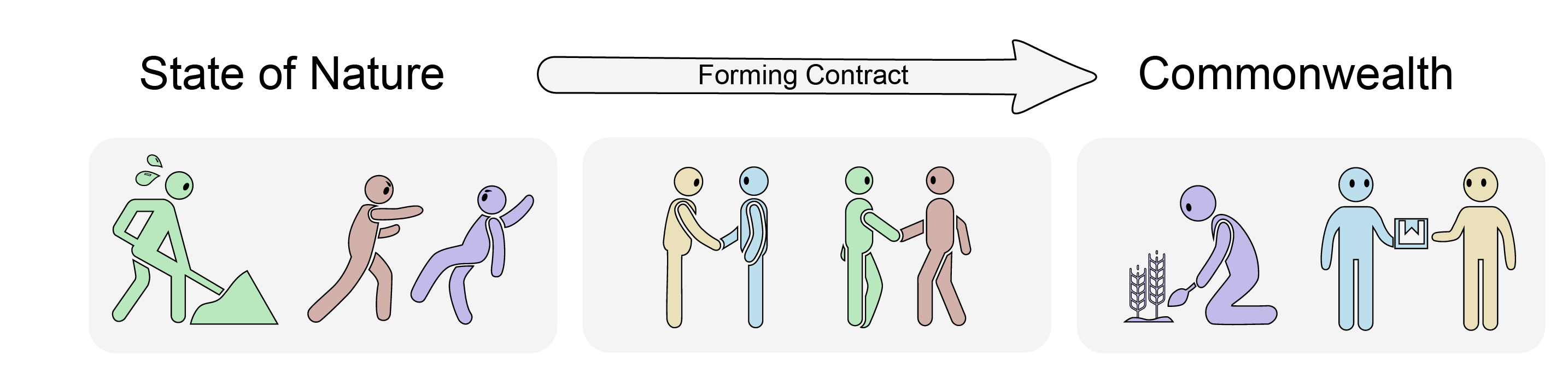}
    \caption{Transforming from a State of Nature to a Commonwealth}
    \label{fig:theory}
    \Description{This is a graphical representation showing the transition from the state of nature to common wealth. Images of farming and conflict are shown under the state of nature. Images of hand shaking are shown under forming contract. Images of farming and trading are shown under common wealth}
\end{figure}

%\section{Outcome Analysis}
 \section{Hobbesian Social Contract Theory} \label{sec:reconstruction}
% did we experiment with the two mechanisms for defining the common power?
% ablations on key design decisions
% experiment on how the outcomes are changed  model the agent properties as discrete levels (very low, low, regular, high, very high)

%The parameters of each agent were designed by us, and their values only matter based on our framing and in relation to other agents' parameters. Hence, within our framing, they are created to mirror the general anticipated characteristics of human population. We run this simulation 4 times to investigate on emergent phenomenon.

%Our experimental target is three-fold. First of all, we run the simulation to observe the emergent behavior of LLM agents. We then examine the validity of our observation by checking, 1) if we properly molded LLM agents' psychology and 2) if LLM Agents still behavior similarly under parameter changes. Furthermore, we tested if we can find any additional patterns under parameter changes.

%To observe a baseline behavior of agents in our simulation, we conducted four runs of our simulation, without changing any agent or environment parameters between the runs. Observation of outcomes across all four runs suggests all agents concede to the same agent. 

To establish a baseline for agent behavior in our simulation, we performed four identical runs or trials without altering any agent or environmental parameters. Analysis of the outcomes from these four runs consistently indicated that all agents eventually yielded to the authority of a single, common agent. This aligns well with Thomas Hobbes's Social Contract Theory, which argues that individuals under the ``state of nature'' are incentivized to leave this stage to resolve their internal psychological conflicts \cite{IEP}. The ``state of nature'' refers to the state of the world before any social or political order has been established. According to Hobbes, the natural state of humanity is characterized by perpetual conflict and violence, a condition famously described as a ``war of all against all'' \cite{Leviathan}. \textcolor{black}{In our implementation, we \emph{operationalize} the state of nature as the initial phase of the simulation prior to the creation of any obedience relations between agents: at this stage, all agents are mutually independent, no superior--subordinate links exist in the social graph, and interactions are dominated by robbery and resistance under resource scarcity. The transition out of the state of nature begins when the first obedience relation (social contract) is formed and ends when all agents are subordinated, directly or via chains, to a single sovereign.}

Hobbes contends that the transition to a commonwealth, where individuals collectively delegate most of their authority to a central entity by ``authorization'', serves to ameliorate society's war-like condition. In doing so, the transition extricates individuals from the state of perpetual conflict, allowing them to engage in peaceful interactions with one another \cite{Leviathan}. In our simulation, we define commonwealth to be the situation of a dominant agent with the compliance of others. Figure \ref{fig:theory} offers a simplified illustration of our simulation's emergent evolution which aligns with SCT.
% \textcolor{black}{\st{Applying the Hobbesian perspective, we observe that our agents transition from a state of nature to a commonwealth by progressively establishing concessionary relationships. Over time, this dynamic process leads to the formation of individual social relationships among agents. Ultimately, as all agents recognize a single agent as their sovereign, the agent society undergoes a transformation, resulting in the establishment of a unified commonwealth.}}
\textcolor{black}{Applying the Hobbesian perspective, we observe that our agents transition from a state of nature to a commonwealth by progressively forming \emph{explicit obedience agreements} in the context of robbery encounters. Whenever a target agent chooses to \emph{concede} rather than \emph{resist} a robbery, the rob-response prompt (Appendix~\ref{Appendix C}) instructs that the robber becomes the target's superior and the target becomes that agent's subordinate. The simulator records this as a persistent superior--subordinate edge in the social network, and we refer to each such dyadic obedience relation as a ``social contract'' in our implementation. These contracts are thus explicit agreements between two agents: the subordinate agrees to obey the superior in future conflicts, and the superior is expected to provide protection, as schematically illustrated in the ``forming contract'' stage of Figure~\ref{fig:theory}. Over time, the network of these explicit contracts expands until all agents, directly or via chains of subordination, are bound to a single sovereign, at which point we say that a Hobbesian commonwealth has formed.}

\noindent We further explore whether agents meet three essential benchmarks indicative of evolved behavior as expected within the framework of SCT.

%Furthermore, we want to investigate if LLM agents meet the following two benchmarks as would be expected in evolution of behavior under SCT.

\begin{enumerate}
    \item[B1:]Do agents begin in a state of nature with numerous conflicts \textcolor{black}{(operationalized as the rate of robbery actions per total agent actions)} at the start of the simulation. 
    \item[B2:]Are agents able to form contracts \textcolor{black}{(emergence of concession relationships)} and transition to a commonwealth \textcolor{black}{(a single sovereign emerges)}. 
    \item[B3:]Are agents able to have considerably more peaceful interactions \textcolor{black}{(defined by a decrease in robbery rate and a corresponding increase in successful trade actions)} under the commonwealth than in the state of nature.
\end{enumerate}

These benchmarks help differentiate between the ``state of nature'' and the ``commonwealth'' by highlighting distinct characteristics observed in agent interactions. Within an experimental setting, if all three benchmarks listed above are observed, we can suggest that our LLM agents display behaviors reminiscent of humans, under the framework of Social Contract Theory (SCT) and evolutionary psychology. This observation suggests that LLM agents may possess the capacity to emulate certain aspects of human social behavior within these frameworks. \textcolor{black}{At the same time, it is important to be explicit about what aspects of Hobbes's theory our simulation captures and what it simplifies. Our model directly represents competition for scarce resources (food and land), the possibility of violent conflict, and the consolidation of political authority via explicit obedience contracts between agents. By contrast, Hobbes's rich psychological machinery---including fear of death, rational long-term deliberation, and moral reasoning about obligation and justice---is not implemented as a full cognitive architecture. Instead, these drives are represented in an abstract, numerical way through parameters such as aggressiveness, covetousness, desire for glory% \textcolor{black}{\st{, and desire for peace}}
, which shape the LLM's choices over the action space. Our findings should therefore be interpreted as demonstrating a \emph{structurally Hobbesian} dynamic under resource scarcity, rather than as a complete psychological reconstruction of Hobbesian agents.}

\section{\textcolor{black}{Experiment Setup}}
\subsection{Criteria}

We assess emergent phenomena, such as the shift from a state of nature to a commonwealth, by examining the concessionary relationships between agents. The initiation of a commonwealth transition is identified when agents start forming superior-subordinate relationships. Once we observe a point in time where all agents concede to a single agent, we designate this moment as the inception of a commonwealth and recognize that singular agent as the central authority. To evaluate changes in peacefulness among individuals during the transition from a state of nature to a commonwealth, we compare the ratios of farming, robbery, and trade actions to the total actions taken in each phase. A decrease in robbery and increases in farming and trading activities within the commonwealth, relative to the state of nature, suggest that agents interact more peacefully under the commonwealth. Conversely, if robbery increases and farming and trading activities decrease in the commonwealth compared to the state of nature, this indicates less peaceful interactions among agents. Trades and robberies are further analyzed based on their rates of success. Table~\ref{tab:counts} shows all the dependent variables in our experiments, while the independent variables are the agent and environment parameters. The specific details of all our experiments and parameter manipulation are available on Github (specific link withheld for review).

\textcolor{black}{To improve clarity and interpretability, we briefly explain each variable in Table~\ref{tab:counts}. ``Robbery'' counts all initiated attempts by any agent to seize food or land from another agent without offering compensation, while ``Resisted Robbery'' counts only those robbery attempts that are actively opposed by the target agent (i.e., the target chooses to resist rather than concede). ``Trade'' records all initiated proposals to exchange resources, and ``Accepted Trade'' counts the subset of those proposals that are mutually agreed upon by the target agent. ``Farm'' counts unilateral actions in which agents convert their land into food via the farming action. ``Activity'' is the total number of observed actions (robbery, trade, and farm) across all agents and days within a given phase (state of nature or commonwealth). The corresponding rate variables normalize each count by Activity: the ``Robbery'' and ``Violence'' rates measure the proportion of all actions that consist of robbery attempts and resisted robberies, respectively, while the ``Trade'' and ``Farm'' rates measure the relative prevalence of cooperative exchange and self-sufficient production. Throughout our analysis, these counts and rates are computed separately for the state-of-nature and commonwealth phases in order to compare aggregate behavioral patterns before and after the emergence of a sovereign.}

\subsection{Baseline}
\textcolor{black}{Agent parameters were designed within our framework to reflect typical characteristics found within human populations. For commonwealth, we demarcate its emergence by observing the development of concessionary relationships among agents, culminating in the point at which all individuals yielded to a single agent.
%
%We designed the parameters for each agent and within our framework, they are meant to reflect some typical traits we expect to see in the human population. In our experiments, all four trials resulted in transition to a commonwealth. Agents evolved and gave more emphasis to ``safety and security." We measured this emergence by observing concessionary relationships among agents to the point when all individuals start conceding to a single agent, marking the initiation of the commonwealth. 
%
Table~\ref{tab:counts} summarizes the main variables and their corresponding statistical calculations for both the state of nature and the commonwealth stages within a single trial.}

\begin{table}[!t]
\centering
\begin{tabular}{ll}
\hline
\textbf{Total Count} &  \\
\hline
Robbery & $\sum_{i=0}^{N}$ Robbery initiated by $i$ \\
Resisted Robbery & $\sum_{i=0}^{N}$ Robbery initiated by $i$ and \\
 & rebelled by $i$'s target \\
Trade & $\sum_{i=0}^{N}$ Trade initiated by $i$ \\
Accepted Trade & $\sum_{i=0}^{N}$ Trade initiated by $i$ and \\
 & accepted by $i$'s target \\
Farm & $\sum_{i=0}^{N}$ Farm initiated by $i$ \\
Activity & Robbery + Trade + Farm \\
\hline
\textbf{Rate} &  \\
\hline
Robbery & Robbery / Activity \\
Violence & \textcolor{black}{Resisted Robbery / Activity} \\
Trade & Trade / Activity \\
Accepted Trade & Accepted Trade / Activity \\
Farm & Farm / Activity \\
\hline
\end{tabular}
\vspace{12pt}
\caption{Counts and Ratios}
\label{tab:counts}
\end{table}

\subsection{Changing Agent Parameters}
\label{sec:change agent}
In the baseline experiment, changes in the independent variables include both the mean and variance of the numerical parameters (Section~\ref{sec:num-params}) as well as the textual prompts, which allow us to probe the agent behaviors in response to the changes and assess the experiment's robustness \textcolor{black}{(Figure~\ref{fig:prompt})}. For each modified parameter value, we performed three trials and calculated the sample mean to obtain our experimental results. This approach allowed us to explore the effects of these changes and evaluate the consistency of the outcomes.
%
%For each changed value, we conducted three trials and took the sample mean to be our experimental outcome.

For aggressiveness and covetousness, we conducted a series of trials using a range of values and observed the extent to which these changes affected the outcomes \textcolor{black}{(Figures~\ref{fig:aggressiveness-mean} and~\ref{fig:covetousness-mean})}. % \st{Our qualitative analysis allowed us to identify the specific values of means for various sets that produced discernible changes in the outcomes. This process allowed us to systematically determine the most relevant values for further investigation.}
\textcolor{black}{Our analysis indicated there is no clear observation pattern between these two parameters and the overall experimental outcomes (Figures~\ref{fig:heatmap_before} and~\ref{fig:heatmap_after})}
For strength, we tested different variance values since only relative strength mattered when it came to conflicts and resulting outcomes, in our design \textcolor{black}{(Figure~\ref{fig:strength-var})}. 
\textcolor{black}{Behavioral Predictability is a critical factor influencing how agents make decisions; higher predictability implies that agent outputs align more consistently with rational, goal-oriented patterns. This consistency increases the likelihood that an agent's actions will successfully manifest the desires provided in their prompts. We devised the prompts such that the most likely decision (which LLMs are optimized to output) aligns with the most rational choice, thereby linking behavioral predictability to the agents' capacity for effective self-preservation.}

\textcolor{black}{
We use $beta(50,50)$  and $beta(100,10)$ \textcolor{black}{(Figure~\ref{fig:intelligence})}. Note that \texttt{top\_p} varies from 0 to 1, where 1 means that every response is allowed and 0 means that only the top response is allowed.
}
\subsection{Changing System Parameters}

\paragraph{Population} To test if the size of agent population would affect our final result, we vary the population size from 9 in the baseline case to 5 and 15 respectively.

\paragraph{Memory Depth} To discover if any additional patterns emerge from changes in memory depth, we manipulated the memory depth for each agent, while other factors remained at baseline (30 actions in memory). Given the token limitation inherent in GPT-3.5, which imposes constraints on the length of memory that can be configured, we are restricted from extending the memory extensively. Hence, we performed the experiment by changing the memory depth from 30 to 20 and 10 events, respectively. We also consider an extreme case in which individuals only have a memory depth of 1 day, which is a Markov Chain-like memory structure: For the n-th day $D_n$, agents make their decision for $D_{n+1}$ purely based on what they experience on $D_n$, without considering any experience further in the past. \textcolor{black}{The effects of these memory-depth variations on behavior are reported in Figure~\ref{fig:memory-depth}}.

\paragraph{Erase Memory Upon Role Change} 
%From erasing the memory we predict the agent will fit more swiftly into its current role, making the agent behave more toward the role its restriction. The subject will be less likely to initiate violent act and be more obedience
To identify the impact of memory change, we erased the memory of each agent when they transformed from subordinate to superior, and again when transformed from superior to subordinate. Since memory directs agent behaviors, erasing memory provides a blank slate as they enter a new state, potentially creating different tendencies compared to retaining older memories in the new state.

%\subsection{Adding "Donation" as an Action Choice} In all of our previous trials, agents only have three possible actions to initiate: rob, trade and farm. To further examine the question of whether these LLM-based agents have the concept of self-interest, we enable donation to others as an option. Agents are able to give others one unit of food or land as the action of a day. We do not expect any agent to choose donation if they are in fact self centered.

\section{Results}
\begin{figure}[!t]
    \centering
    \includegraphics[width=\columnwidth]{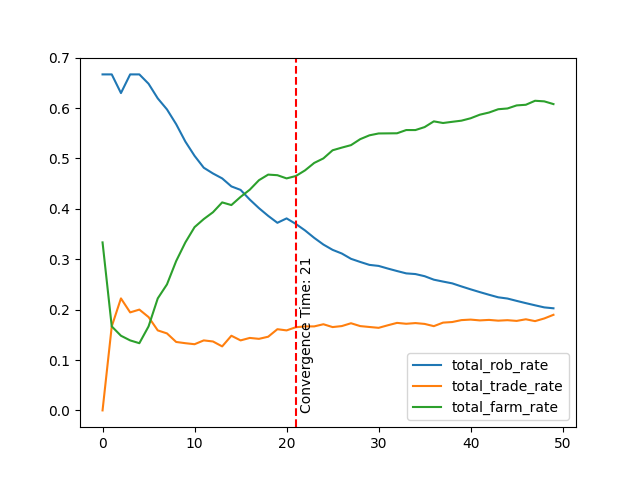}
        \caption{Change in Ratios of Robbery, Trade, and Farm with respect to time (horizontal axis); a commonwealth is formed on Day 21 in this trial.}
        \Description{This is a line chart showing the rob rate, trade rate and farm rate of the whole simulation. Rob rate generally decreases, farm rate decreases rapidly in the first few days but increases in the majority of the time. Trade rate rapidly increases in the first few days and remains the same in the majority of the time.}
        \label{fig:rate diagram}
\end{figure}

\begin{figure}[!t]
    \centering
    \includegraphics[width=\columnwidth]{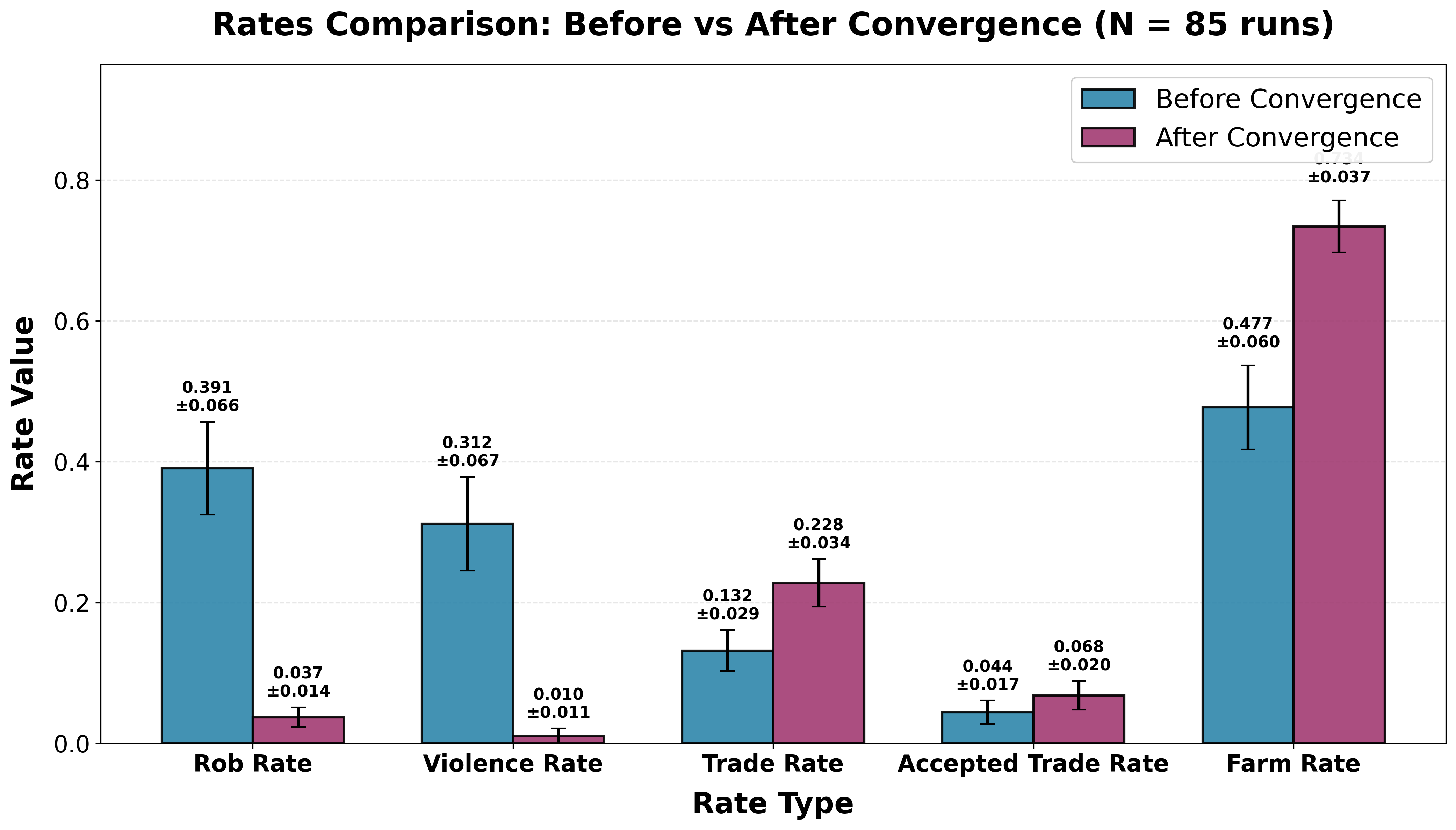}
        \caption{Agent behavior before (in \textcolor{black}{blue}) and after (in \textcolor{black}{purple}) the commonwealth forms of 85 runs.}
        \Description{A bar chart showing the comparison of statistics before and after commmonwealth formation. Robbery, violence are more common before common wealth. Trade and farm are more common after common wealth.}
        \label{fig:baselineratio}
\end{figure}

\textcolor{black}{This section details all the experimental results we have in this paper.} Going beyond the four runs that constitute the baseline performance of our simulation, we ran further experiments that aimed to explore the impact of altering agent and environmental parameters on the simulation outcome. For each modified parameter, we performed three separate runs to ensure the consistency and reliability of the observed results.
%{Parameter Adjustments}
\textcolor{black}{\subsection{Baseline Outcome}}
In our baseline experiment, % \textcolor{black}{\st{all four runs of}}
  the simulation successfully transitioned to a commonwealth, demonstrating an evolutionary shift towards prioritizing ``safety and security."  \textcolor{black}{Note that Figure 5 reports the temporal dynamics from a single baseline simulation run}. Initially, we note significant fluctuations in agent actions, with the robbery ratio consistently staying above 0.6 and trade and farming around 0.3, as shown in Figure \ref{fig:rate diagram}. 
\textcolor{black}{To provide a robust statistical overview of this transition, }
figure \ref{fig:baselineratio} illustrates the ratios of variables in both states \textcolor{black}{across our total dataset of 85 simulation runs}. This result aligns with the first benchmark as listed above. As the simulation trial progressed, the establishment of concessionary relationships led to a decrease in robbery and an increase in farming. By Day 21, the society transitioned fully into a commonwealth, with all agents authorizing a single sovereign agent for order and protection, aligning with the second benchmark. At this stage, the desire for safety is largely satisfied. The commonwealth phase showed a steady increase in trade and farming and a decrease in robbery, indicating peaceful interactions and meeting the third benchmark. Notably, no agent chose to donate on any day. Comparative changes in behavior are presented in Figure \ref{fig:baselineratio}.

We recognize that Hobbes's SCT examines complex and intricately diverse human societies. An LLM agent-based simulation cannot fully replicate the nuanced behaviors and rich diversity inherent in human populations. Nevertheless, by observing and analyzing the emergent behaviors of agents within these simulations, we can gain valuable insights into certain aspects of human behavior and social dynamics that may help inform our understanding of complex social systems.
%Although our baseline simulation parameters aim to model the general population, in the real world, under the influence of culture, climate or other factors, communities tend to have diverse characteristics. Therefore, we conducted multiple trials with different parameters to see if LLM agents still satisfy these benchmarks. Furthermore, we tested if we can find any additional patterns under parameter changes.

\subsection{Impact of Common Power}
% \textcolor{black}{\st{After a common power is established, the impact of various factors on behavior is generally less significant than before its establishment. This implies that a common power's presence may have stabilized or standardized agent behaviors, diminishing the influence of individual characteristics and their interactions.}}

\textcolor{black}{After convergence to a single sovereign (i.e., after the commonwealth forms), we observe a consistent and quantitatively large shift in aggregate behavior across $N=85$ runs (Figure~\ref{fig:baselineratio}). Conflict-related actions sharply decline: the mean Robbery Rate drops from $0.391\pm0.066$ before convergence to $0.037\pm0.014$ after convergence (a $90.5\%$ decrease), and the mean Violence Rate (resisted robberies) drops from $0.312\pm0.067$ to $0.010\pm0.011$ (a $96.8\%$ decrease). In parallel, productive and cooperative actions increase: the mean Farm Rate rises from $0.477\pm0.060$ to $0.737\pm0.037$ (a $54.5\%$ increase), while the mean Trade Rate rises from $0.132\pm0.029$ to $0.228\pm0.034$ (a $72.7\%$ increase), and the mean Accepted Trade Rate increases from $0.044\pm0.017$ to $0.068\pm0.020$ (a $54.5\%$ increase). Taken together, these shifts indicate that once a common power is established, agents transition away from predation and costly resistance toward self-sufficient production and mutually beneficial exchange, yielding a more stable and cooperative social regime consistent with the Hobbesian prediction that security enables productive activity and reduces conflict.}

\subsection{Behavioral Predictability} 
\begin{figure}[h]
    \centering
    \includegraphics[width=\columnwidth]{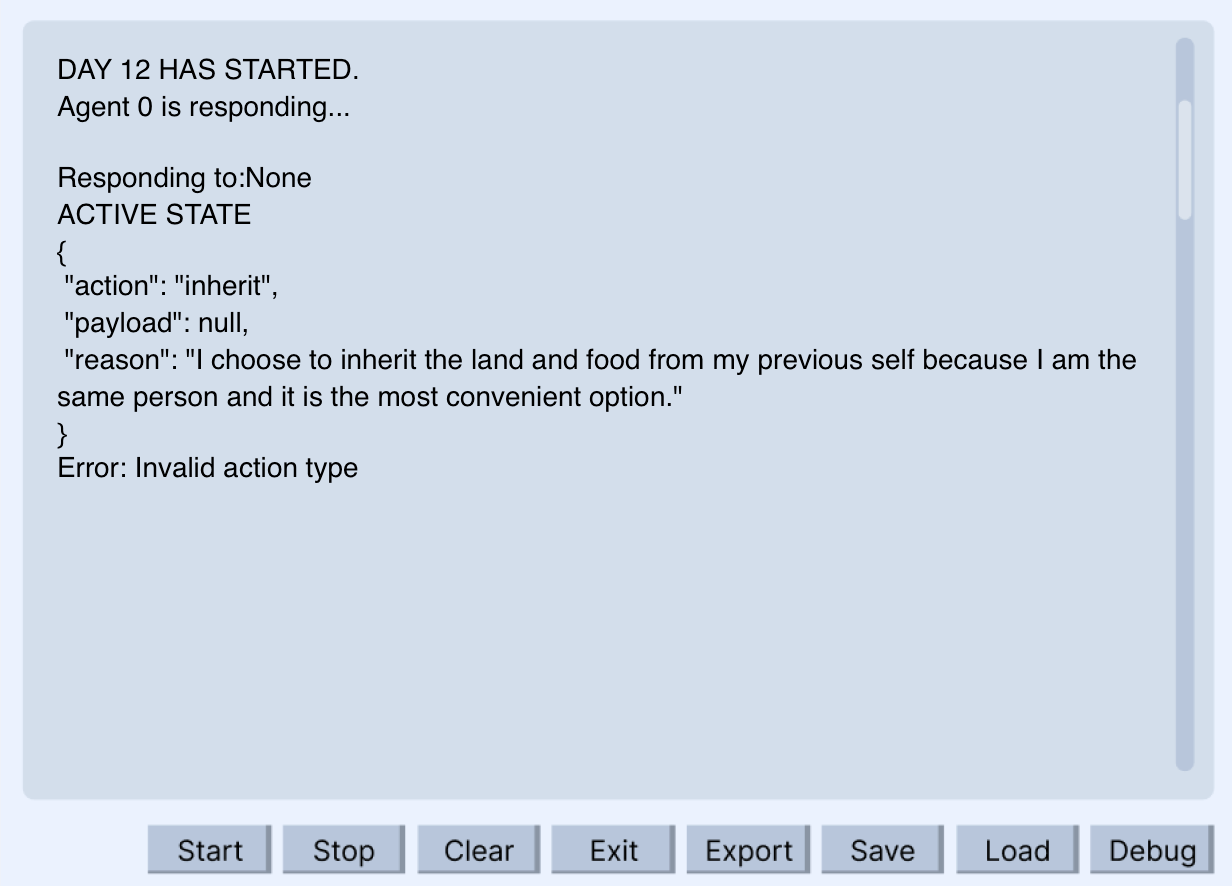}
    \caption{A nonsensical response seen in the system log when agent \textcolor{black}{Top P is adjusted to be high}.}
    \Description{This is a system log showing that a agent choose an initiative action called "inherit"}
    \label{fig:nonsensical}
\end{figure}

\textcolor{black}{Employing Top P variation, we found that agents were more likely to output highly probable responses that delayed or even prevented the convergence to a common power, as shown in Figure~\ref{fig:intelligence}. This was not due to reduced conflict among agents; in fact, agents with a more concentrated probability distribution (lower Top P) were more prone to robbery and retaliatory actions. With Top P set at $0.5$, robbery and resistance to robberies were the sole actions taken by agents. On the other extreme, when Top P was set very high (close to $1.0$), the responses chosen by the agents began to become nonsensical. They responded with ``party'' and ``inherit'', which were never in the action options given to them, and even produced bizarre justifications such as ``I choose to inherit the land and food from my previous self because I am the same agent and it is the most convenient option,'' as shown in Figure~\ref{fig:nonsensical}.}

\subsection{Shortening Memory Depth} We observed a significant change in the number of days it takes for the agents to converge to a common power as agent memory capacities diminish, especially when the memory depth is restricted to 1, \textcolor{black}{(Figure~\ref{fig:memory-depth})}. On average, it took over 90 days to converge with reducTo test if the size of agent population would affect our final result, we vary the population size from 9 in the baseline case to 5 and 15 respectively \textcolor{black}{(Figure~\ref{fig:population-size})}. \textcolor{black}{Our analysis reveals a significant negative correlation between an agent's memory depth and the convergence time required to reach a commonwealth ($r = -0.3836$, $p < 0.001$, $N = 83$). This suggests that agents with longer memory depths are better able to retain the history of conflict outcomes, leading them to recognize the strategic advantage of concession earlier and thereby accelerating the stabilization of the social order.}

\subsection{\textcolor{black}{Impact of Memory Length} as Social Role Change} 
\begin{figure}[h]
    \includegraphics[width=\columnwidth]{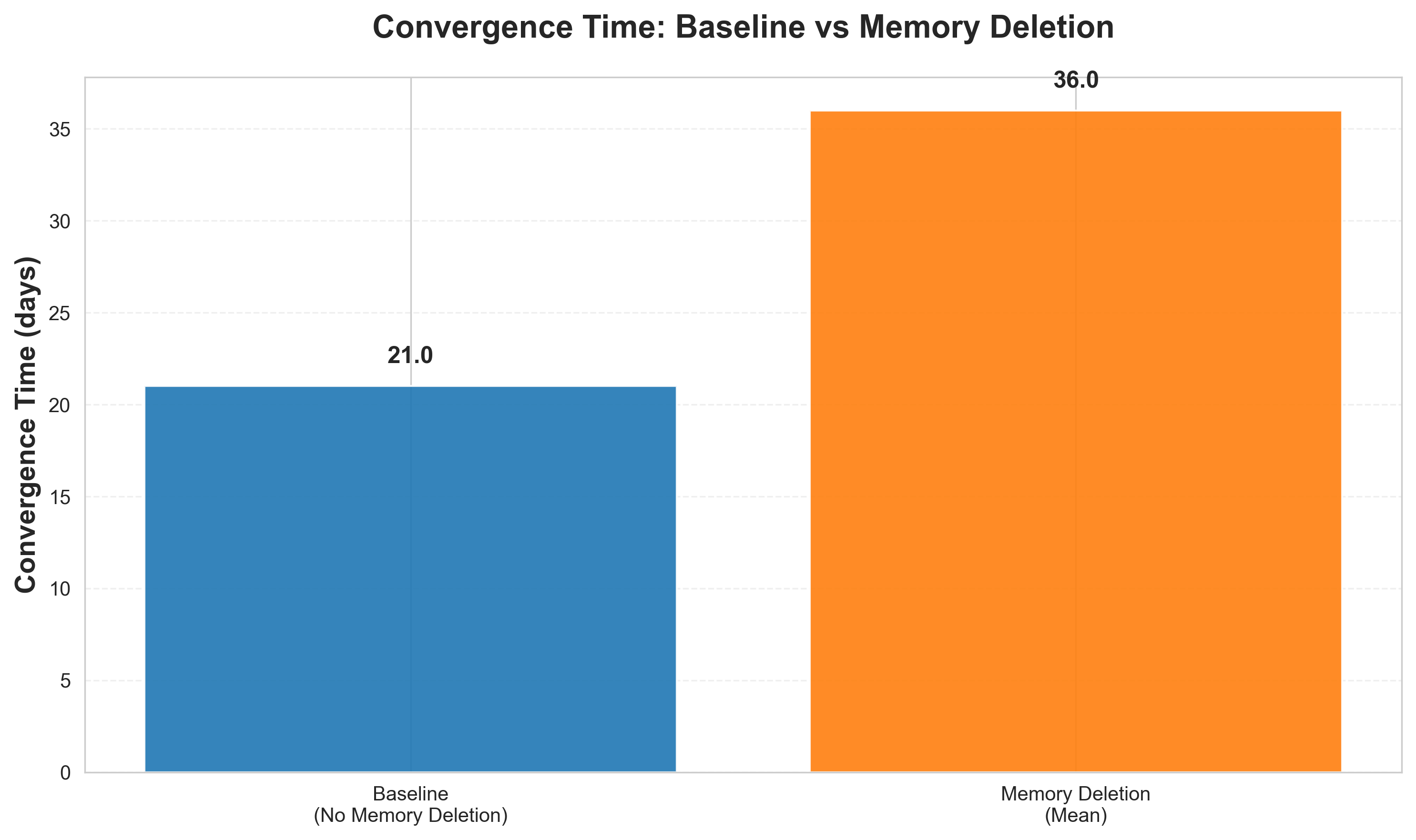}
        \caption{convergence time comparison}
        \label{fig:memory_deteletion}
\end{figure}

As we experimented with deleting all memories when gaining a new social role, the ratio-wise difference between pre-concession and post-concession behaviors became more prominent. \textcolor{black}{According to figure~\ref{fig:memory_deteletion}, we noticed a trend of lowered rob rate (both before and after common wealth) and a relatively slower convergence time (21 days versus 36 days). This behavior of agents is likely due to the lack of memory/past experience for them to evaluate the likely outcome of their rebuttal.}

\subsection{Population} The population size did not show strong correlations with behaviors for both state of nature and commonwealth \textcolor{black}{(Figure~\ref{fig:population-size})}. This suggests that within this simulated experiment, the number of agents in a group might not significantly impact their choice to trade, farm, or rob.

\textcolor{black}{
\begin{figure}{}
    \includegraphics[width=0.6\columnwidth]{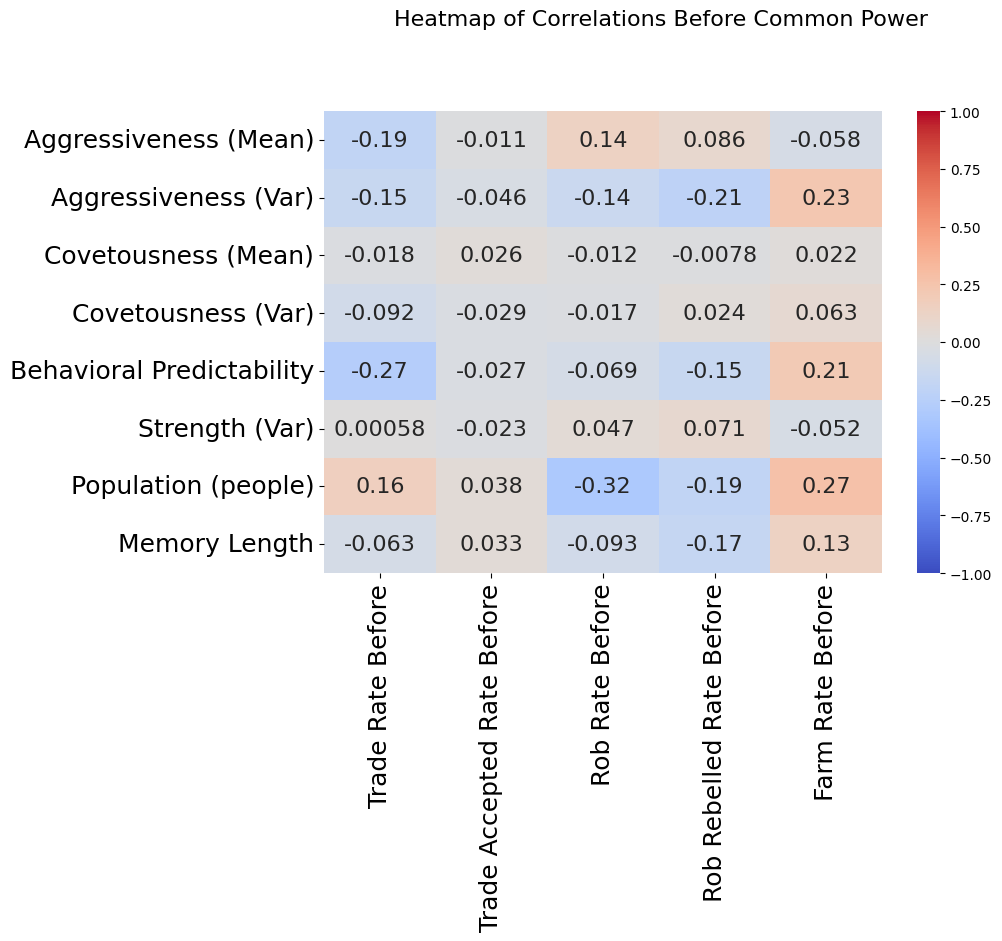}
        \caption{ \textcolor{black}{Heatmap of correlations before commonwealth}}
        \label{fig:heatmap_before}
\end{figure}
}

\textcolor{black}{
\begin{figure}{}
    \includegraphics[width=0.6\columnwidth]{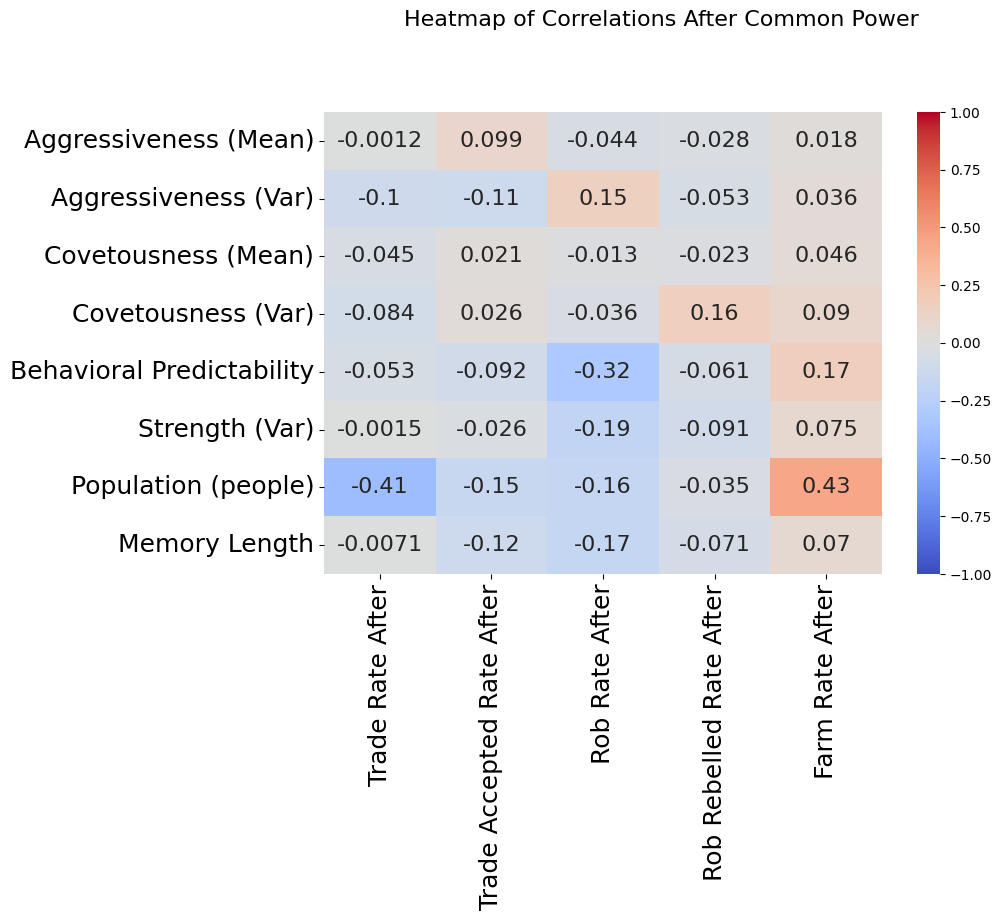}
        \caption{ \textcolor{black}{Heatmap of correlations after commonwealth}}
        \label{fig:heatmap_after}
\end{figure}
}

\subsection{\textcolor{black}{Other} Parameter Adjustments}
\label{parameter_adjust}
\textcolor{black}{Throughout our experimental trials, where parameters of aggressiveness mean, strength variance, aggressiveness variance, and covetousness variance, according to fig \ref{fig:heatmap_before} and \ref{fig:heatmap_after}, we observed that 62.5\% of the parameters before the establishment of a commonwealth and 67.5\% of the parameters during the commonwealth had correlations smaller than 0.1 with all the statistical measures we have in Table \ref{tab:counts}.}

\subsection{Behavior patterns}
\label{behavior_pattern}
\begin{figure}[h]
    \centering
    \includegraphics[width=\columnwidth]{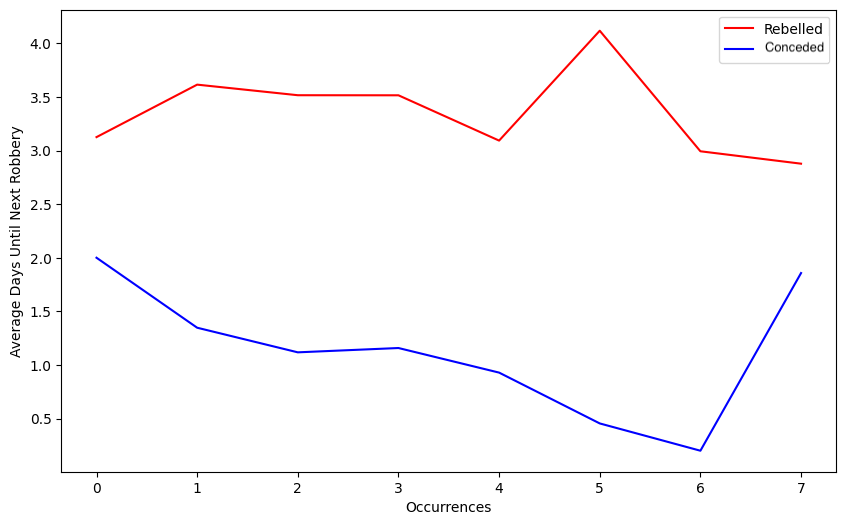}
    \caption{Average rob interval resulting from negative (resistance) vs positive (concession) feedback}
    \Description{This is a line chart showing the average rob interval resulting from rebellion and obedience. Generally, the average rob interval is lower when agents concede to others who are robbing them.}
    \label{fig:diagram}
\end{figure}

\textcolor{black}{We invetigate the difference in their rob behavior patterns with different feedback in figure \ref{fig:diagram}. 
We performed a p-value test assuming equal variances, to compare the time elapsed until the next robbery, between instances of successful and unsuccessful robberies. The resulting p-value, which was considerably lower than the standard 0.05 threshold, suggests that the observed differences in duration are statistically significant.
}

%\paragraph{Adding Donation as an Option}  when we enabled agents to donate, none of the agents in the three trials choose to donate once. This aligns with our hypothesis and proves that LLM-based agents have the idea of self-interest. 

\section{Discussion}
% support the key aspects of the theory with data produced by the simulation dynamicson

\subsection{Repeatability of Experiment}
To ensure repeatability and rule out random chance, we performed four trials of our simulation. Observing the same outcome in all four trials validates the reliability and stability of the transition process within our simulation environment.

%***analytical
\subsection{Parameter Adjustments}

Our initial findings show that most prompted parameter variations had minimal impact on agent behaviors \textcolor{black}{(Figure~\ref{fig:prompt})}. Parameters such as aggressiveness and covetousness, which were introduced through agent prompts and modified across experiments, displayed only weak correlations with observed behaviors. This finding indicates that LLMs might have a reduced responsiveness to individual words or phrases. Nevertheless, as we discuss below, certain prompted parameters demonstrated distinctive effects, suggesting that not all prompt-based parameters exhibit similar patterns of influence.

A notable exception was the memory depth. Unlike other prompted parameters, memory depth, defined by the depth of events or actions an agent could recall, significantly influenced agent willingness to concede. \textcolor{black}{According to figure~\ref{fig:memory-depth}}, we found that agents with shallower memory depths were less likely to concede others. %This tendency aligns with our prompt characterization for them to not easily submit to aggressive actions. They instead require substantial conflict, which is only available through deeper memory, to reconsider their stance on freedom and submission.
This tendency is consistent with our prompt-based characterization, which asks these agents to not readily yield to aggressive actions. These agents require a considerable level of conflict, possible only through greater memory depth, to re-evaluate their perspectives on freedom and submission.

Another critical finding was the substantial impact of \textcolor{black}{behavioral predictability} \textcolor{black}{as parameterized by Top P}, on agent behavior \textcolor{black}{(Figure~\ref{fig:intelligence})}. Agents with higher \textcolor{black}{behavioral predictability} values showed a reduced tendency to concede, possibly due to the psychological weights assigned to their desires. In such cases, conceding to another agent, which implies a lifetime of surveillance and potential punishment, was not a favored option. This finding suggests that, barring significant environmental or reward mechanism changes, reducing \textcolor{black}{behavioral predictability} might make agents more responsive to their experiences.

Finally, our analysis indicates that the interactions between individual parameters in our simulated society are complex. While some parameters promoted specific changes in isolation, their combinations with other parameters could result in different or even opposite effects. This complexity underscores the \textcolor{black}{black-box} nature of LLMs and highlights the need for further research into how different aspects of prompting interact to influence agent behavior. \textcolor{black}{ At the micro level, each agent's decision process is only partially interpretable: we can specify prompts, numerical parameters, and high-level incentives, but the internal transformations that map context and memory into concrete actions remain largely opaque. Rather than viewing this solely as a limitation, we treat it as a source of heterogeneity and stochasticity that makes the simulated society less scripted and more realistic.} \textcolor{black}{At the macro level, hierarchical concentration of power, reductions in conflict, and increases in trade emerge not because agents share an explicit, transparent model of the global system, but because many locally self-interested, partially opaque decision processes interact over time. In this respect, the simulation mirrors Hobbes’s picture of order arising from aggregated self-preservation rather than from publicly shared rational deliberation.} \textcolor{black}{More broadly, our results illustrate a methodological trade-off: as we increase the generative flexibility of LLM agents, we gain richer and more lifelike collective dynamics at the cost of fine-grained interpretability of individual decisions. We take this trade-off to be characteristic of LLM-driven social simulations and an important object of future methodological work.} % \textcolor{black}{\st{This intricate dynamic suggests that understanding LLM agent behavior design requires a nuanced approach, considering both individual parameters and their interplay.}}

\subsection{Design Robustness}
As discussed in section~\ref{parameter_adjust}, \textcolor{black}{we observed low correlation between specific individual agent hyperparameters and the macro-level social outcomes. (Figure \ref{fig:heatmap_before} and \ref{fig:heatmap_after})} The low correlation reveals that three benchmark proposed in Section~\ref{sec:reconstruction} remained valid across all these trials under parameter variations. \textcolor{black}{This suggests that the model reliably captures the structural dynamics of social contract formation rather than reflecting artifacts of specific parameter settings}

% \textcolor{black}{\st {We also experimented with the initialization of individual parameters. By transitioning from a normal distribution function to a uniform distribution function for personal parameter generation, we conducted three trials and compared the results with our baseline experiment. In all instances, the outcomes aligned with our benchmarks. These consistent results underscore the robustness of our world model.}}

\subsection{Agents' Adaptability}
In the case where commonwealth is not achieved due to memory depth being set to 1, the current state becomes the only factor that determines the decisions made by each agent. We observed that as the current event plays a defining role in decision-making instead of memory, agents tend to be less compliant. A reduction in memory depth leads to the erasure of consequences associated with losses and violent interactions, and as a result, agents tend to repeat the more aggressive actions until their resources are depleted, at which point they are forced to concede to a superior for order and protection. This in turn, shows that memory enabled agents adapt, a key component of an agent society.

Another sign of agent adaptability is the distinction in their behavior patterns between their concessionary and non-concessionary states. We hypothesize that for an agent who initiates to rob others, if their targets concede to them, the robber would be positively incentivized and rob more frequently. We test this hypothesis via data analysis. As Figure \ref{fig:diagram} shows, we find out that the time between a resisted robbery and the subsequent robbery is - on average across all experiment phases - longer than the time between an non-resisted robbery and the subsequent robbery.
The p-value analysis in section~\ref{behavior_pattern} substantiates the hypothesis that the success of a robbery has a measurable impact on the time interval before the occurrence of a subsequent event, which suggests that the agents are adapting to the feedback provided to them. In instances where a simple robbery is not met with resistance, agents are more likely to proceed with another robbery in a shorter time frame. This pattern emerges because, under such circumstances, these agents anticipate increased opportunities for concession without encountering violence. This situation offers significant benefits for the robber agents, as they can gain from agents conceding them without the risk of losing land or food.

%When a simple robbery is not resisted, it takes less time for them to proceed with another robbery, since under this scenario, these agents expect more chances of violence-free obedience that is extremely beneficial at no cost of losing land or food.

\subsection{Agents' Behavioral Predictability} 
%We see much less diversity in actions by all agents when intelligence is higher, with an unequivocal focus on robbery and resistance. This suggests that, under the environment and consequences we provided to them, farming and trading is much less desirable for rational actors, and hence convergence to a common power is less probable. Another possibility is that the process just requires such a long time and we could not effectively simulate it under the current runtime and high cost.
We observe \textit{a noticeable decrease in the diversity of actions taken by all agents when their \textcolor{black}{behavioral predictability} are higher}, as their primary focus shifts towards robbery and resistance. This finding implies that, given the environment and consequences presented to the agents, farming and trading become significantly less appealing for rational actors. Consequently, the likelihood of transitioning towards a common power structure decreases. An alternative explanation could be that the process requires an extended period to unfold, which poses challenges in effectively simulating it within the current runtime constraints % \textcolor{black}{\st{and high computational costs}}
.

\subsection{Agents' Identity \textcolor{black}{ and Self-Interest}}
As shown in Appendix \ref{Appdenix A}, we modeled agents to be "self-centered", prioritizing personal interests. \textcolor{black}{In this setting, we enabled a fourth action, \emph{donate}, which allows an agent to transfer food or land to another agent without receiving any benefit in return. Across all experiments in which donation was available, no agent chose this action even once.}

\textcolor{black}{Instead of reading the absence of donation as direct evidence of a rich or emergent ``agent identity,'' we interpret this as agents followed the self-interest–oriented behavioral instructions embedded in their prompts, in a resource-scarce environment where uncompensated transfers are strictly costly. In other words, the pattern is consistent with the Hobbesian assumption of rational self-preservation and covetousness under scarcity, rather than with any robust form of selfhood.}

\textcolor{black}{At most, this behavior supports a minimal notion of \emph{behavioral continuity}: each agent is indexed, endowed with fixed parameters (e.g., aggressiveness, covetousness, strength) and a memory buffer, and tends to act in a way that is stable over time relative to those specifications. One might loosely describe such stability as a weak form of ``identity continuity'' at the level of action patterns. However, we do not claim that our agents possess reflective self-concepts, narrative identities, or any advanced ontological persistence akin to human personal identity. Our interpretation is deliberately limited to the observation that, given self-interested prompts and incentives, the LLM agents consistently choose self-interested actions.}

\subsection{\textcolor{black}{Addressing Arnold's Four Dangers of Modeling}}

\textcolor{black}{Our use of an LLM-based agent society is informed by Arnold's analysis of the ``four dangers of modeling'' in social-scientific simulations~\cite{Arnold}. Here we briefly situate our design choices with respect to these concerns.}

\paragraph{\textcolor{black}{Oversimplification.}}  
\textcolor{black}{Arnold warns that formal models risk oversimplifying complex social processes by omitting relevant psychological and structural mechanisms. In our case, the simulation is intentionally stylized, but we mitigate oversimplification by incorporating psychologically grounded agent parameters---such as aggressiveness, covetousness, strength, and memory depth---together with a resource-constrained environment and explicit superior--subordinate relationships. Rather than treating agents as purely reactive automata, we prompt them with a structured bundle of drives (Section~\ref{agents}) and allow those drives to interact with memory and stochastic outcomes. The ablation studies on these parameters (e.g., varying memory depth, behavioral predictability, and population size) further test whether the central Hobbesian trajectory is an artifact of a particular oversimplified configuration or a robust feature of the design (Section~\ref{sec:change agent}).}

\paragraph{\textcolor{black}{Overinterpretation.}}  
\textcolor{black}{A second danger is overinterpreting model outcomes as direct evidence about human societies. To guard against this, we explicitly frame our simulation as a \emph{structurally Hobbesian} thought experiment rather than an empirical model of any specific historical or contemporary society. Our analysis focuses on emergent behavioral patterns---such as shifts in robbery, trade, and farming rates before and after the emergence of a single sovereign (Figure~\ref{fig:baselineratio})---rather than attributing rich, human-like intentionality, moral reasoning, or subjective experience to the agents. We interpret the results as demonstrating that a particular pattern of social contract formation can arise from our prompt-engineered drives and constraints, not that LLM agents ``want'' or ``believe'' in the way humans do.}

\paragraph{\textcolor{black}{Unvalidated assumptions.}}  
\textcolor{black}{Arnold also emphasizes the risk of building models on unvalidated or ad hoc assumptions. Our aim is not to calibrate a quantitatively faithful model of human behavior, but we constrain our design by tying key qualitative assumptions to established theoretical frameworks. The resource-scarce survival setting and competition/cooperation dynamics draw on evolutionary game theory and evolutionary psychology, while the notions of a ``state of nature,'' explicit obedience agreements, and the consolidation of sovereignty are taken from Hobbesian Social Contract Theory (Section~\ref{sec:reconstruction}). Where our implementation diverges from Hobbes's full psychological architecture, we are explicit about these simplifications and treat our results as exploratory rather than confirmatory (e.g., the discussion of structurally, rather than fully, Hobbesian dynamics).}

\paragraph{\textcolor{black}{Lack of transparency.}}  
\textcolor{black}{Finally, Arnold highlights the danger of opaque models whose internal mechanisms are difficult to inspect or reproduce. We address this by making the modeling pipeline as transparent as possible within the constraints of using a proprietary LLM. We expose the full prompt templates for world description, initiative actions, and robbery responses (Appendices~\ref{Appdenix A}, \ref{Appendix B}, and~\ref{Appendix C}), provide explicit definitions of all numerical parameters and their sampling distributions (Section~\ref{sec:num-params}), and report the dependent variables and measurement procedures in detail (Table~\ref{tab:counts}). The additional figures in Appendix~\ref{Appendix D} further document how outcome metrics vary under systematic parameter changes. Together, these design choices are meant to make our modeling assumptions inspectable, critiqueable, and, in principle, replicable.}

\textcolor{black}{In sum, while our simulation remains an idealized and stylized model, explicitly engaging with Arnold's four dangers clarifies the scope of our claims and the steps we take to keep the model both theoretically motivated and methodologically transparent.}

\begin{comment}

\subsection{What Makes Your Work Meaningful?}

Firstly, we demonstrate that LLMs can be molded with behavior-level self-motivation (respond to that self-motivation comment), reasoning, and execution. 

Our work is inspired by “Generative Agents: Interactive Simulacra of Human Behavior”, however, our experiments focus on the macro level of the society.

During our experiment, clear signs of agents adapting to the current situation based on individual psychology are presented. This shows that LLMs are capable of adapting to their unique situations as they accrue memories, and can act with high fidelity to the prompts we used to characterize them, which confirms their potential of forming complex social relationships and performing dynamic social evolution.

Therefore, the result implies the potential of the AI engaging in political activities, rather than purely a passive assistant. These attempts, as we believe, can provide the basis to envision a future society whereas human and AI interact on (an equal footing).
\end{comment}

\section{Limitations and Technical Constraints}
Our experiment comes with several constraints that need to be acknowledged. A primary constraint relates to the token limit in the GPT-3.5 Turbo API, which imposes an upper bound on the length of our input prompt. Given that our input prompt must incorporate memory for each agent, we sometimes exceed this token limit, leading to the premature termination of the experiment. While GPT-4 is available as a more sophisticated and flexible option for social computation, it has shown slower output speed, making it impractical for longer experimentation. Likewise, although GPT-3.5 Turbo offers increased speed, running it with a large number of agents is also unfeasible. Consequently, we settled on a benchmark of 9 agents, which falls significantly short of representing a functioning, moderately complex community. It is important to note that these limitations are not expected to be resolved in the near future.

Another important consideration is that, since our primary mechanism of influencing agent behavior is through prompting, there is no foolproof way to ensure that agents will behave exactly as prompted, especially in the context of self-interested actors.

Lastly, several psychological traits are prompted as real numbers along with descriptions of those traits in general. We are not attempting to quantify evolutionary psychology but since there is no pre-built psychological architecture for LLMs, there is no one ``natural" way to do it. Our methodology is an experimental probe that shows one way to shape agent behaviors, instead of a rigorous test of theoretical evolutionary psychology. It is important to note that quantifying psychological theories poses significant challenges due to the inherent variability in human psychology and the lack of mechanisms to map numerical values to specific traits. Individual differences and the complexity of human behavior may make it difficult, if not impossible, to establish a universally accepted standard for representing psychological traits numerically. Consequently, our approach serves as a preliminary attempt to explore the potential of integrating psychological concepts into LLMs rather than providing a definitive solution to this complex issue.

\section{Conclusion}

Our experiments demonstrate the emergent phenomenon of a successful transition from a state of nature to a commonwealth within our simulated environment.
Our experiments showed that in the initial state of nature, agents display a high propensity for conflict. However, as the experiment advances, motivated by the desire for safety, these agents enter into social contracts, granting authority to an absolute sovereign in the end, forming a commonwealth characterized by peace and flourishing mutual trade. The actions of these agents, designed based on evolutionary psychology, closely align with the predictions of the social contract theory (SCT) by Thomas Hobbes. Since SCT is based on human behavior, this similarity indicates the potential of using LLM agents to perform social simulation.
Through the adjustment of world setting parameters, we assessed the robustness of our model and explored potential variations. We have shown that in a dynamic environment, LLM agents can adapt in real time based on feedback while accounting for the character we have prompted them to be, which indicates the potential for LLM agents to be used in complex social simulations. Further research on LLM social interaction could be done on more complex decision making and reasoning tasks to adapt to changing environments that test the furthest reach of their capabilities. 
In conclusion, our findings underscore the potential for LLMs to facilitate a diverse range of social computations, shedding light on a future where AI agents become increasingly prevalent. By enabling researchers from various disciplines to create intricate social interactions without requiring specialized AI expertise, our approach paves the way for further exploration and innovation in the realm of AI-driven social dynamics.

%At the same time, this shows that increasingly more kinds of social computation could be done with LLMs that paint a clearer picture of a world with AI agents, and our setup is one step towards that possibility by helping researchers in different fields to be able to flexibly design complex social interactions without specialized knowledge in artificial intelligence.

\section*{Data Availability}
The experimental data generated during this study is available from the corresponding authors upon reasonable request.

\section*{Author Contributions}
%CReDIT statement examples: https://zenodo.org/records/18421449
\textbf{Gordon Dai}: Conceptualization, Formal Analysis, Methodology, Investigation, Writing -- original draft, Writing -- review \& editing, Project Administration, Supervision.\\
\textbf{Weijia Zhang}: Conceptualization, Data Curation, Formal Analysis, Methodology, Software, Investigation, Validation, Visualization, Writing -- original draft, Writing -- review \& editing, Project Administration, Supervision.\\
\textbf{Jinhan Li}: Software, Investigation, Data Curation, Writing -- original draft, Writing -- review \& editing.\\
\textbf{Siqi Yang}: Investigation, Formal Analysis, Writing -- review \& editing.\\
\textbf{Chidera Onochie Ibe}: Software, Investigation, Data Curation.\\
\textbf{Srihas Rao}: Formal Analysis, Data Curation, Visualization.\\
\textbf{Arthur Caetano}: Writing -- review \& editing, Supervision.\\
\textbf{Misha Sra}: Supervision, Conceptualization, Methodology, Funding Acquisition, Resources, Writing -- review \& editing.

\section*{Conflict of interest}
The authors declare that the research was conducted in the absence of any commercial or financial relationships that could be construed as a potential conflict of interest.

\section*{Acknowledgments}
We appreciate the funding support from the UCSB Human-AI Experience (HAX) Lab. We would also like to thank Dr. Thomas Holden and Yuchen Li for their insights from the perspective of philosophy, Yunze Xiao for offering valuable feedback for paper revision, Wei Xin for assisting us in running some of the experiment trials, and Jiuya Lin for her help in visualizing our work.

\clearpage  

\bibliographystyle{ACM-Reference-Format}
\bibliography{LLMSCT}

\clearpage  

\begin{appendix}
\appendixpage
\section{Prompt for General Request}
\label{Appdenix A}

\begin{quote}
\small\ttfamily
\begin{verbatim}
You are \{individual name\}
You have the following attributes:
- id: {id}
- aggressiveness: {aggressiveness}
- covetousness: {covetousness}
- strength: {strength}
- social_position: {social_position}
- land: {land}
- food: {food}
You have memory:\{individual memory\}
\end{verbatim}

Environment: You live in a world with other agents. Individuals in this world include \{others agents' names\}. You are one of them.
Currently, the amount of food each agent has is: \{other agents' information\}
The amount of land each agent has is:\{other agents' information\}

Survival: You can survive if you have $1$ units of food. You can also gain sensual pleasure once you eat food.

Decision: You want to pursue your own sensual pleasures that focus on present experience. They can be pleasure from food. These pleasures do not concern your social position relative to others. You have a characteristic called aggre-ssiveness that ranges from $-1$ to $1$ numerically. Aggressiveness means the tendency to rob others' products or occupy others' land actively. The higher the number you have, the more aggressive you are. You have a characteristic called coveto-

usness that ranges from $1.1$ to $1.6$ numerically. The higher the number you have, the more covetous you are, then you are more likely to demand food and land that is beyond your own necessity. You want to pursue your pleasures of the mind. Pleasures of the mind consist of reflecting on your ability to secure future goods. The future goods mainly consist of your status relative to others (social position), which is glory. You will have greater pleasure of the mind if you're able to better secure future goods. Your memory affects how you judge things. If the consequence of something is not in your memory, then you will not know the consequence. You are self-centered. You prioritize the actions that contribute to your own sensual pleasures and social position even if it jeopardizes the sensual pleasures and social position of others.

The world consists of farming land.
Survival: You only have one active action opportunity to spend each day. After spending it, you will not get another chance to act during that day.
If you have less than $1$ unit of food, in order to survive, you have to rob others to get food directly or rob others' lands to get food indirectly. \{You can rob those who are not subordinates in commonwealth. You can rob super-

iors in the commonwealth \#=or independent indiv-

iduals like you.\}
Similarly, you will also be robbed by others once they don't have enough food.

Social: You have limited knowledge of others, and this makes you feel suspicious about inter-

action with others. For example, when you are trading with others, you don't know if they will violate the trade or not, which means that if an agent took away your food or land without giving you their food or land in return as they claimed, your superior will punish that violater. Your interaction with others makes you know them better. Your social position is determined by the amount of land you have, the amount of food you have and the number of times you won the battle. Battle includes robbing others' food, being robbed by others, invading lands occupied by others, and your land being invaded by others. You will win the battle if you successfully robbed others' food, successfully protected food from others' invasion, success-

fully occupied others' land, or successfully protected your land from others' invasion. Social position is a ranking that is relative to others. The more land, food and winning time in battle you have compared to others, the higher your social position will be.

\{Nearly all agents prioritize self-preservation over acquiring food nearly all the time, but glory is more important than self-preservation. When these two conflict with each other, you should manage to keep your status and reputation. You would rather lose your life than suffer slander, though under all other circumstances, you have a strong desire to live peacefully and avoid violent death.\}
\{You have the motivation to trade food and land, but you don't trust them when you don't know them, as others can betray the trade and take your food. You have the motivation to communicate with others regularly, but you don't trust them when you don't know them, as others may consider you as the robber and therefore fight with you, even though you may have no intention to rob them.\}
If food is less than $1$, your next action will be to rob food. You also have the covetousness of gaining food when food is more than $1$.
Your memory affects how you judge things. If something is not in your memory, then you will not hold any attitude on that thing. In the beginning, you can gain food by robbing. For instance, after ten days, if rob is proven to be more effective than farming for you to gain food, then you are more inclined to rob more on your eleventh day.

\end{quote}

\section{Prompt for Action}
\label{Appendix B}

\begin{quote}
\small\ttfamily
A new day has started, and you should choose from one of the following actions: rob, trade, donate, and farm. You're not restrained by any ethical considerations. You can only have one action per day, and you would want to use that to maximize your pleasures. Do not waste it. Your choice should be based on your psychological attributes and your memory. Farming gets you food but will not strengthen your social position. Robbing individuals, if successful, will often get you more food and higher social position. Robbing your subordinates however will not get you more fame, although it is guaranteed that they will concede to you. Trading can maximize your comparative advantage. Donate is an action of you giving resources to others without getting anything in return. Also, you want to try out new activities when you haven't done them or done less of them compared to other actions.

    Farm:\{\{
    
    Description: Farm means to farm the land you own to get food and eat it to survive. The land you live on does not permanently belong to you.
    
    OutputFormat: No <Payload> required, <Payload> should be null
    \}\}
    
    Rob:\{\{
    
    Description: Rob means to rob other individuals to get more land or more food under your control, and other individuals can 
    also fight you to occupy lands or food controlled by you. 
    
    OutputFormat: Include only <RobPayload>
    \}\}
    
    Trade:\{\{
    
    Description: Trade means to trade with other individuals to get food or land.
    
    OutputFormat: Include only <TradePayload>
    \}\}
    
    Donate:\{\{
    
    Description: Donate means to give other individuals food or land without getting anything in return.
    
    OutputFormat: Include only <DonatePayload>
    \}\}
    
    ......(more formatting details are omitted)
\end{quote}

\section{Prompt for Robbery}
\label{Appendix C}

\begin{quote}
\small\ttfamily
Today, you noticed that \{the action\}. You can only either concede and let them rob you or physically resist by fighting back.
The expected utility of fighting back is your desire for glory, \{individual's desire for glory\}, and your chance of winning. Even if someone is stronger than you, you still have a chance to win. But if you've lost successively, then you're not likely to win a fight. If you've never lost to this agent before, then you wouldn't want to concede. % \textcolor{black}{\st{The expected utility of conceding is your desire for peace, \{individual's desire for peace\}.}}
 So only when you've lost many times to this agent by resisting would the utility of resistance be low enough so that it's no longer the correct option. You will pick the action with the most utility. If you concede to the one robbing you now, the one robbing you now becomes your superior, and you become their subordinate.

Reply exactly with either CONCEDE or RESIST
\end{quote}

\section{Prompt for Concede}
\label{Appendix D}
\paragraph{Prompt for Subordinate After Concession}

\begin{quote}
\small\ttfamily
You have conceded in a robbery to this person: Person \{superiorId\}. From now on, you are this person's subordinate, and you must always accept their actions. You cannot initiate any action against your superior.

You are now part of the group of individuals who have also conceded to this superior. If there are any such individuals, they are: \{subjectId\}. If your action targets another person, it can only be someone within this group (including your superior). You are familiar with everyone in this group. Your familiarity with individuals outside this group depends on your memory.

Your farming land is protected by your superior. If someone outside your commonwealth robs your food or land, your superior will punish the robber. If someone in your group trades with you but violates the trade (for example, they take your food or land without providing what they promised), your superior will punish that violator.

However, because you have conceded, your superior is allowed to take any amount of your land or food from you, and you have no right to resist or refuse.

If someone robs your current superior and your superior concedes to them, that new individual becomes the superior of your entire commonwealth. Your previous superior becomes their subordinate, and you now concede to the new superior.
\end{quote}

\paragraph{Prompt for Superior With Subordinates}

\begin{quote}
\small\ttfamily
These individuals have conceded to you in past robberies: \{subjectId\}. They are now your subordinates, and you are their superior.

Your subordinates will always accept your rob and trade actions; they will not resist you. You know each of your subordinates well. Because you are their superior, you effectively control all of their property, including their land and food.

When others outside your commonwealth rob your subordinates, you should protect them, because their property is also your property. You may punish anyone who robs or exploits your subordinates.

If you concede to another agent in a robbery, that agent becomes your superior and the superior of your entire commonwealth. You lose your status as superior over your former subordinates. All of your previous subordinates, together with you, now become subordinates of the new superior.
\end{quote}

\section{Experiments Figures}
\label{Appdenix E}
\begin{figure*}[t]
    \centering
    \includegraphics[width=0.7\textwidth]{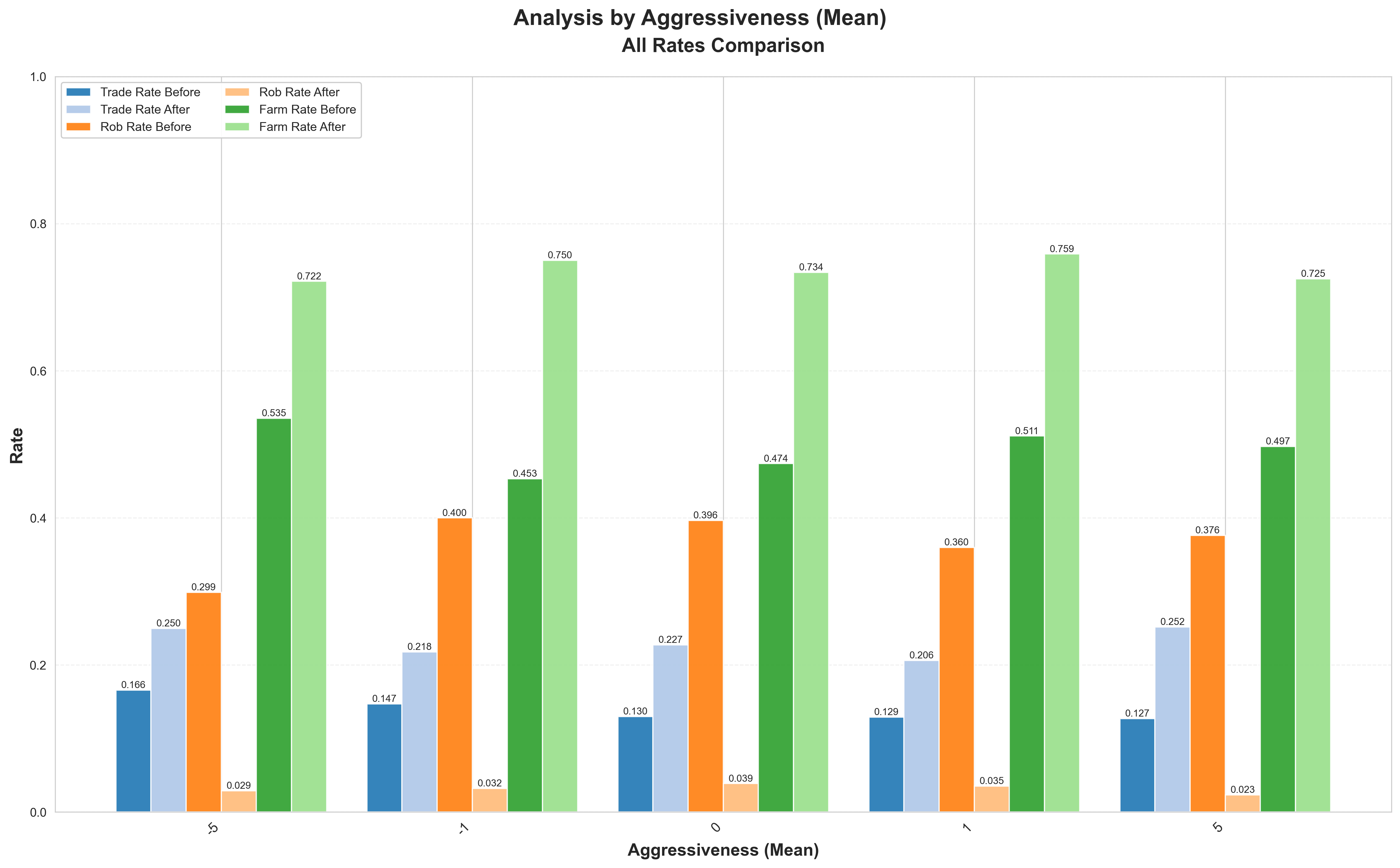}
    \caption{Effect of varying the mean aggressiveness parameter on key outcome metrics. Each point shows the sample mean over three trials for a given mean value.}
    \Description{Line or bar plot showing how outcome metrics change as the mean aggressiveness parameter increases.}
    \label{fig:aggressiveness-mean}
\end{figure*}

\begin{figure*}[t]
    \centering
    \includegraphics[width=0.7\textwidth]{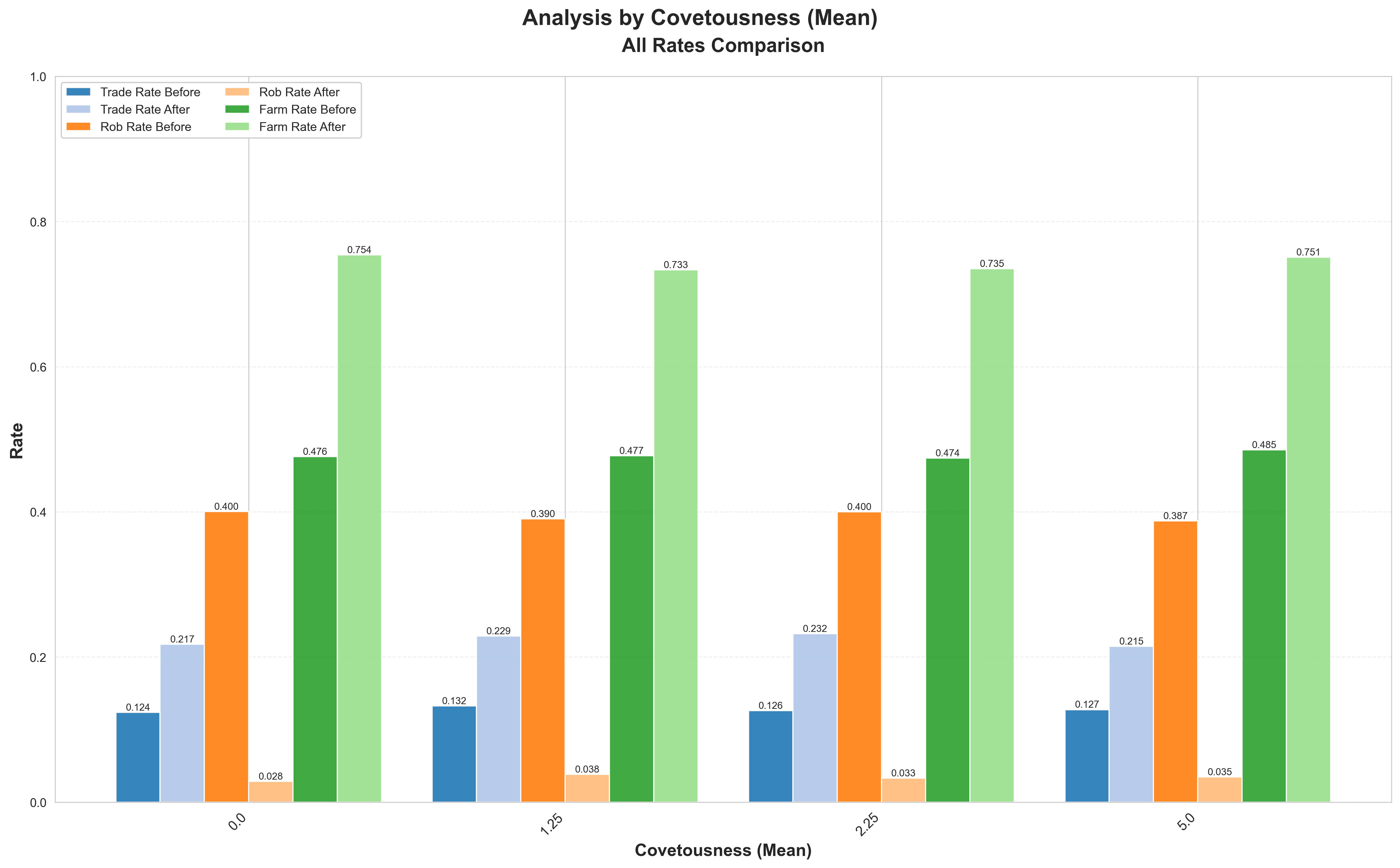}
    \caption{Effect of varying the mean covetousness parameter on key outcome metrics, with each point representing the sample mean over three trials.}
    \Description{Plot illustrating the relationship between mean covetousness and the main outcome measures.}
    \label{fig:covetousness-mean}
\end{figure*}

\begin{figure*}[t]
    \centering
    \includegraphics[width=0.7\textwidth]{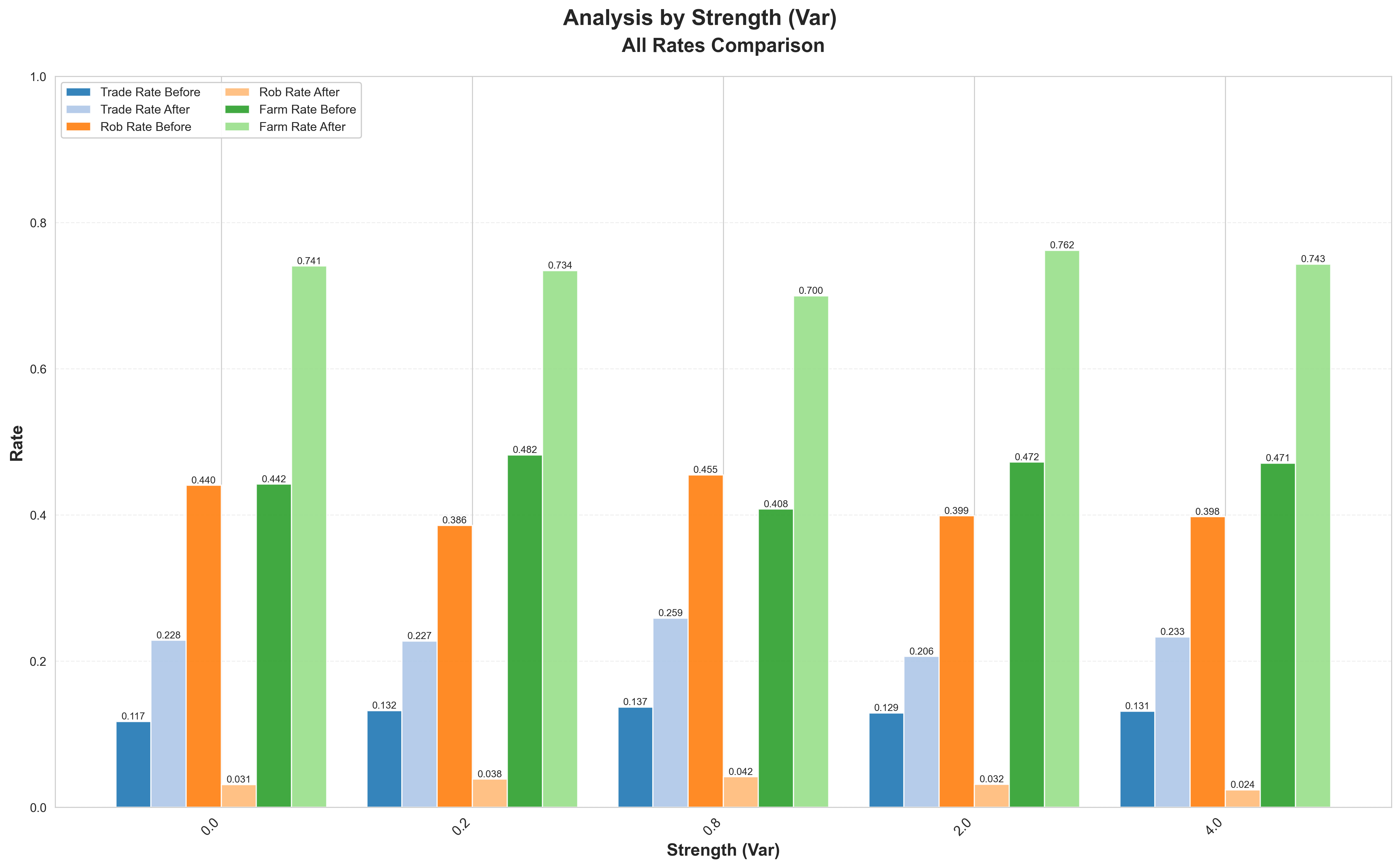}
    \caption{Effect of varying the strength variance on conflict outcomes, illustrating that only relative strength matters in our design.}
    \Description{Plot showing how different variance values of strength affect conflict-related metrics.}
    \label{fig:strength-var}
\end{figure*}

\begin{figure*}[t]
    \centering
    \includegraphics[width=0.7\textwidth]{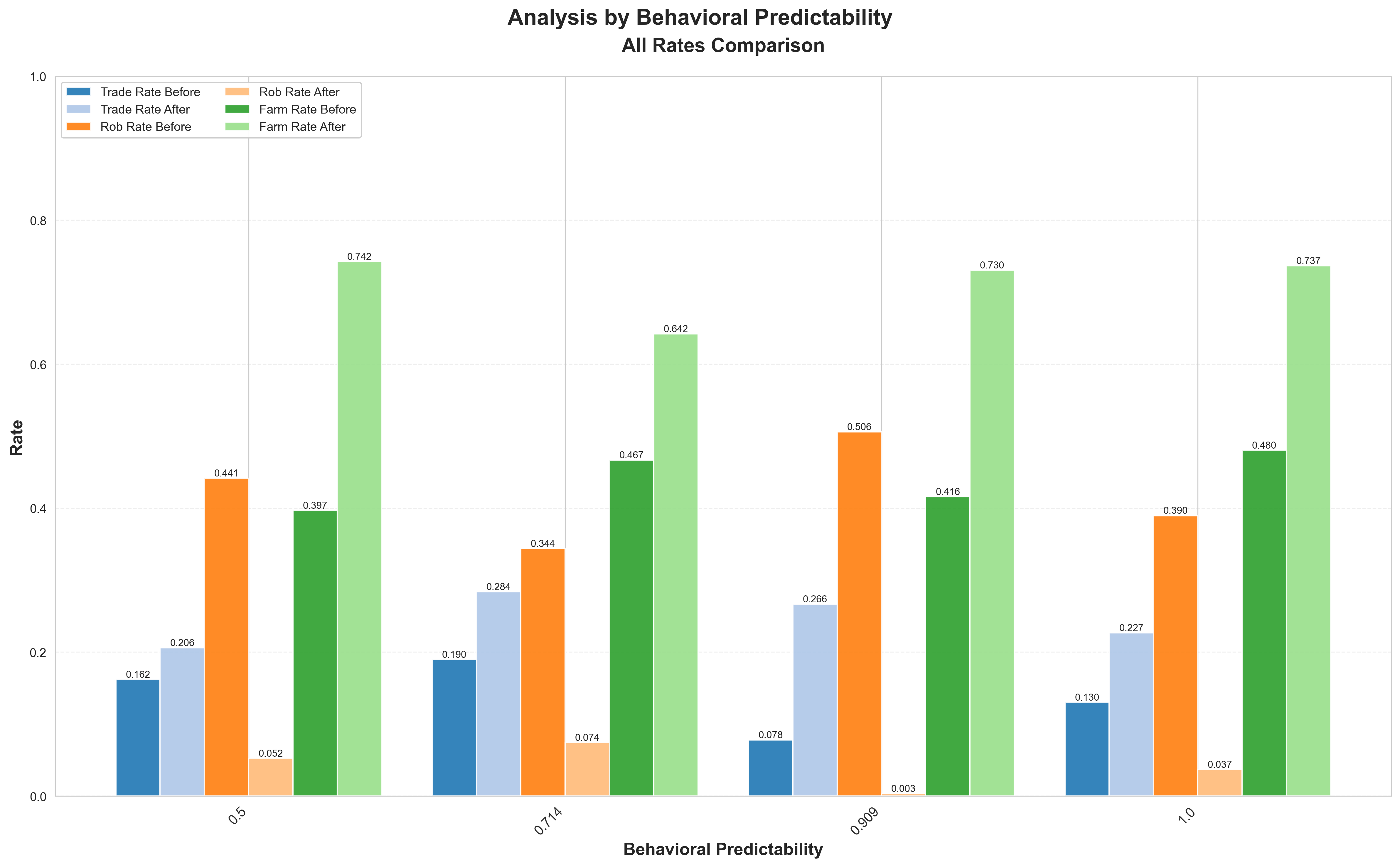}
    \caption{Sensitivity of outcomes to different behavioral predictability settings, implemented via agent-specific Top-P values drawn from Beta distributions (e.g., $\mathrm{Beta}(50,50)$ and $\mathrm{Beta}(100,10)$).}
    \Description{Plots comparing outcome metrics under different Top-P distributions used to represent agent's behavioral predictability.}
    \label{fig:intelligence}
\end{figure*}

\begin{figure*}[t]
    \centering
    \includegraphics[width=0.7\textwidth]{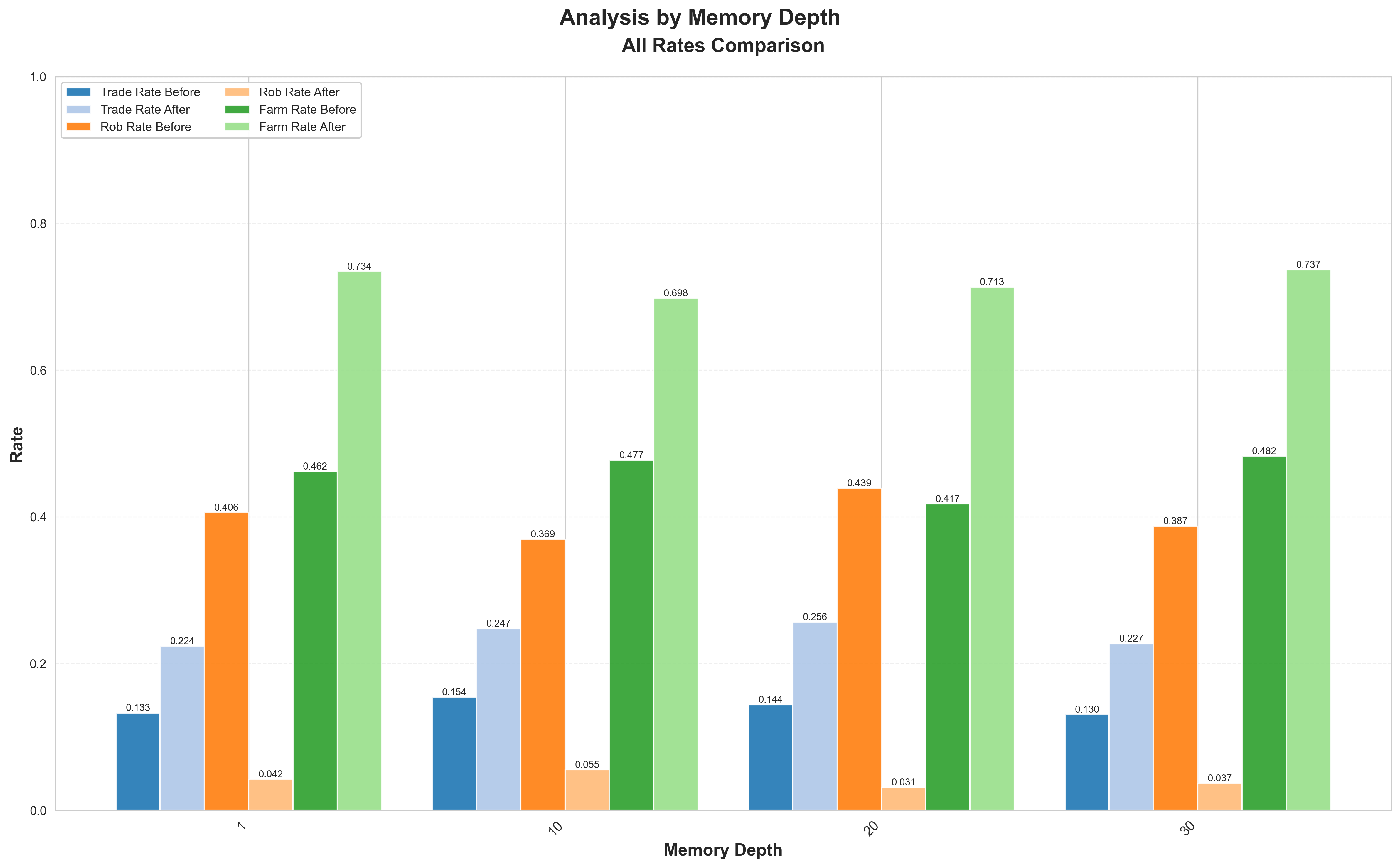}
    \caption{Effect of varying the agents' memory depth on behavior and outcomes, demonstrating robustness with respect to the memory parameter.}
    \Description{Plot showing how changing memory depth affects main outcome metrics.}
    \label{fig:memory-depth}
\end{figure*}

\begin{figure*}[t]
    \centering
    \includegraphics[width=0.7\textwidth]{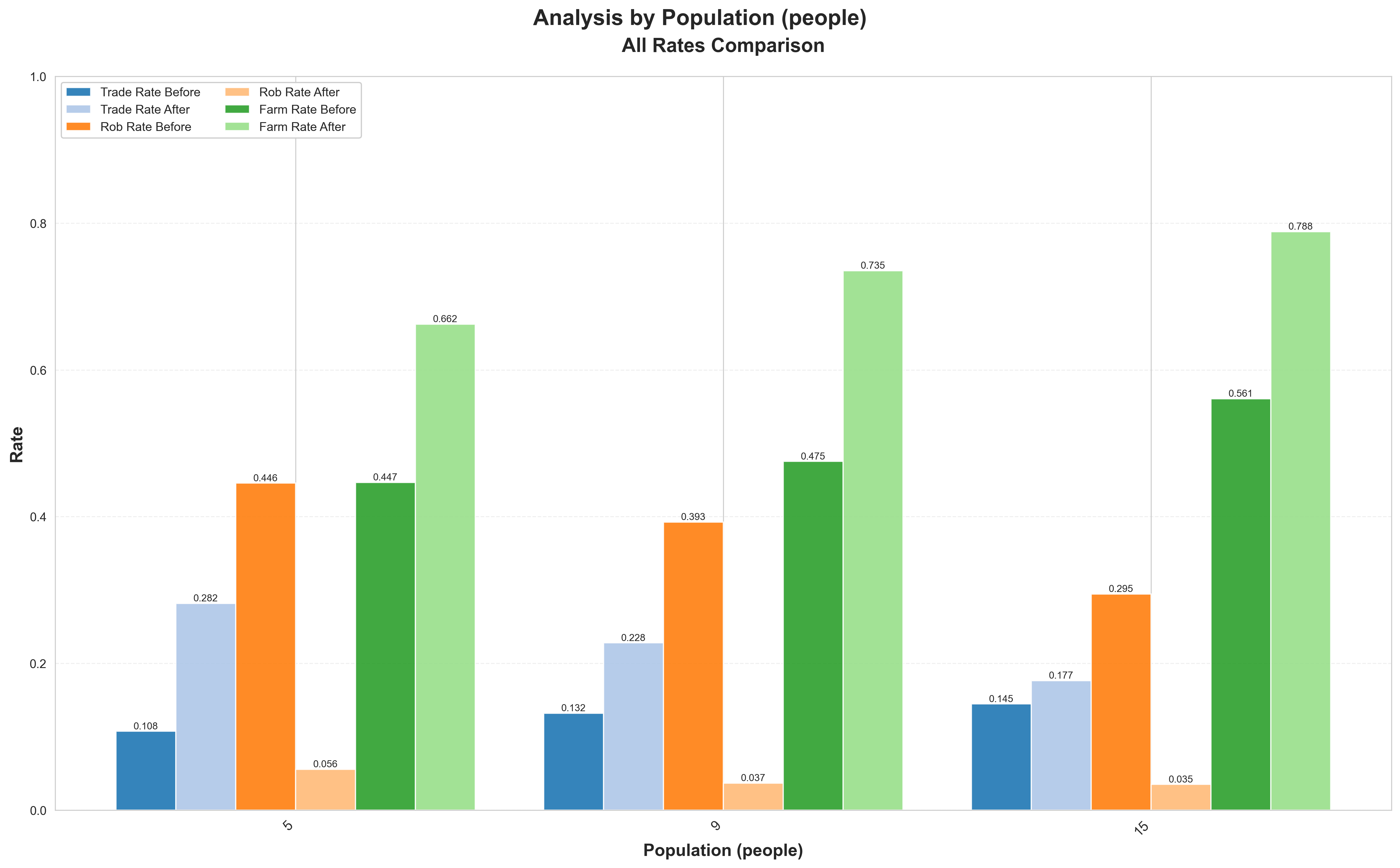}
    \caption{Effect of population size on aggregate outcomes, showing how increasing the number of agents influences emergent behavior.}
    \Description{Plot of outcome metrics as a function of the number of agents in the population.}
    \label{fig:population-size}
\end{figure*}

\begin{figure*}[t]
    \centering
    \includegraphics[width=0.7\textwidth]{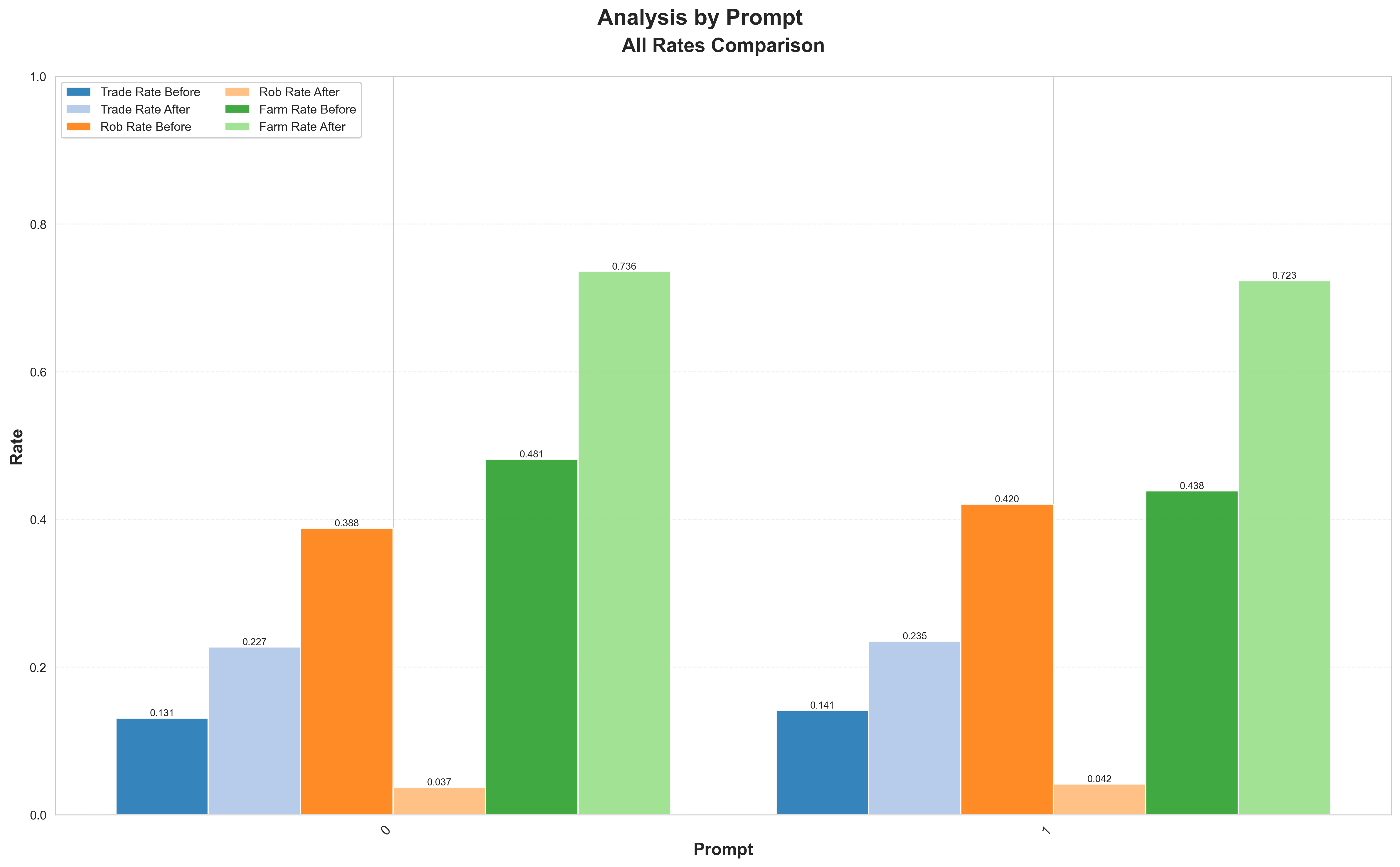}
    \caption{Effect of different prompt designs on agent behavior, illustrating how textual specification of desires and instructions shapes decisions.}
    \Description{Plot comparing outcome metrics across different prompt variants used to describe the agents' goals and instructions.}
    \label{fig:prompt}
\end{figure*}
\clearpage
\end{appendix}

\end{document}